\UseRawInputEncoding
\RequirePackage{fix-cm}
\documentclass{svjour3}         %onecolumn(standard format)
\smartqed  % flush right qed marks, e.g. at end of proof
\usepackage{graphicx}
\usepackage{epstopdf}
\usepackage{geometry}
\usepackage[numbers,sort&compress]{natbib}

\usepackage{bm}
\usepackage{amsmath,amssymb,amsfonts}
\usepackage{algorithm}
\usepackage[noend]{algpseudocode}
\usepackage{graphicx}
\usepackage{multirow}
\usepackage{etoolbox}
\usepackage{verbatim}
\usepackage{lineno}
\usepackage{setspace}
\usepackage{textcomp}
\usepackage[tight]{subfigure}
\usepackage{caption}
\usepackage{subfig}
\usepackage{epstopdf}
\usepackage[misc]{ifsym} % tongxunzuozhetubiao

\begin{document}
\title{Dunhuang murals contour generation network based on convolution and self-attention fusion %\thanks{Grants or other notes
}


\author{Baokai Liu\textsuperscript{1}         \and
        Fengjie He\textsuperscript{2}        \and
        Shiqiang Du\textsuperscript{1,3}      \and
        Kaiwu Zhang\textsuperscript{1}         \and
        Jianhua Wang\textsuperscript{1}
             %etc.,~\Letter
}

\institute{%
	\begin{itemize}
		\item[\textsuperscript{\Letter}] {Shiqiang~Du } \\
		%		Tel.: +123-45-678910 \\
		%		Fax: +123-45-678910 \\
		\email{shiqiangdu@hotmail.com}
		\at
		\item[\textsuperscript{1}] Key Laboratory of China's Ethnic Languages and Information Technology of Ministry of Education, Chinese National Information Technology Research Institute, Northwest Minzu University, Lanzhou, Gansu, 730030 China
		\at
		\item[\textsuperscript{2}] China Mobile Design Institute Co., Ltd. Shaanxi Branch, Xi'an, Shaanxi, 710000 China
      	\at
		\item[\textsuperscript{3}] College of Mathematics and Computer Science, Northwest Minzu University, Lanzhou, Gansu, 730030 China
	\end{itemize}
}

%\institute{F. Author \at
 %             first address \\
  %            Tel.: +123-45-678910\\
   %           Fax: +123-45-678910\\
    %          \email{fauthor@example.com}           %  \\
     %      \and
      %     S. Author \at
       %       second address
%}

\date{Received: date / Accepted: date}

%\twocolumn[
%\begin{@twocolumnfalse}
\maketitle
\begin{abstract}
Dunhuang murals are a collection of Chinese style and national style, forming an autonomous Chinese-style Buddhist art. It has a very great historical and cultural value and an important research value.
Among them, the lines of Dunhuang murals are extremely general and expressive. It reflects the character's distinct character and intricate inner emotions.
Hence, the outline drawing of the murals is of great importance in search of the Dunhuang culture.
The generation of contours of Dunhuang murals belongs to the detection of image contours, which is an important branch of computer vision, aimed at extract salient contour information from the images.
Although convolution-based deep learning networks have achieved good results in extracting image contours by exploring the contextual and semantic features of images. However, as the receptive field broadens, some detailed local information is lost.
As a result, it is impossible for them to generate reasonable outline drawings of murals.
In this paper, we propose a novel edge detector based on self-attention combined with convolution to generate line drawings of Dunhuang murals.
Compared with existing edge detection methods, firstly, a new residual self-attention and convolution mixed module(Ramix)
is  proposed to merge local and global features into feature maps. Secondly, a novel densely connected backbone extraction network is designed to efficiently propagate information rich in edge features from shallow layers into deep layers.
It is apparent from the experimental results that the method of this paper can generate richer and sharper edge maps on several standard edge detection datasets. In addition, tests on the Dunhuang mural dataset show that our method can achieve very competitive performance.
\keywords{ Edge detection \and  Ramix \and Line drawings  \and Dunhuang murals}
\end{abstract}

%\end{@twocolumnfalse}
%]

\section{Introduction}
\label{sec:Introduction}
Dunhuang murals have a long and brilliant artistic achievement. They are an important part of the history of the development of Chinese painting, the most dazzling brilliant pearl in traditional Chinese culture, and the remarkable representative of human civilization.
In order to protect these precious cultural heritages,
many museums digitize murals into painting images, and use appropriate algorithms to realize image restoration.
In the process of manually repairing damaged murals, some highly skilled painters first sketch the outlines of these murals on paper, and then restore the original appearance of the murals by coloring. This work will require a lot of manpower and material resources. Therefore, the generation of digital mural line drawing has become a very important task in mural restoration work \cite{liu2022dunhuang}.
However, there are two challenges that make it difficult for existing edge detection methods to obtain relatively clear outlines of murals:

$\textbf{Challenge 1:}$ The lines and outlines of Dunhuang murals are hick and uneven, with light and dark changes, varied colors, and complex intersections. In the existing edge extraction network based on convolution, with the increase of network depth and receptive field, some local feature information is lost, which makes some important edge details in the mural contour map lost.

$\textbf{Challenge 2:}$ The Dunhuang murals are damaged in local areas. The edge extraction network based on the convolution cannot make full use of the context information and local feature information, so the generated mural contour map has many lines with abnormal edges.

The line drawing task of digital murals  falls under the category of edge detection, and the purpose of edge detection is to accurately extract the contour area of the object. It has a very wide range of applications in image segmentation, object detection and image inpainting. In edge detection tasks, traditional methods use low-level features of images (color, brightness and texture) to capture edges \cite{marr1980theory,martin2004learning,tariq2021tumor,mohamed2021flexible}. Since these methods ignore the semantic information of the image, the generated edge map has serious false positive phenomenon. The deep learning-based convolutional neural network has made significant progress in the field of edge detection due to its powerful semantic extraction ability. But for convolutional neural networks, as the receptive field expands, some important details are inevitably lost gradually. To solve this problem, most convolutional neural networks adopt the method of fusion of deep and shallow features \cite{ziou1998edge,dollar2006supervised}.
However, due to the limited receptive field of the convolutional network, it cannot capture the long-distance semantic relationship above, which makes the generated edge map have a certain amount of noise.
In recent years,
due to the excellent performance of the vision transformer in capturing long dependencies, an edge detection method EDTER based on transformer is proposed \cite{pu2022edter}. This method utilizes the context information and local information of the whole image to obtain clear edges and achieves very good results. But this approach is computationally expensive. Inspired by \cite{pan2022integration,srinivas2021bottleneck}, we integrate a self-attention mechanism and a convolutional fusion module with dense networks. The model effectively improves the performance of edge detection without significantly increasing the amount of computation.

\begin{figure}[t]%
	\centering
	\includegraphics[width=0.8\textwidth]{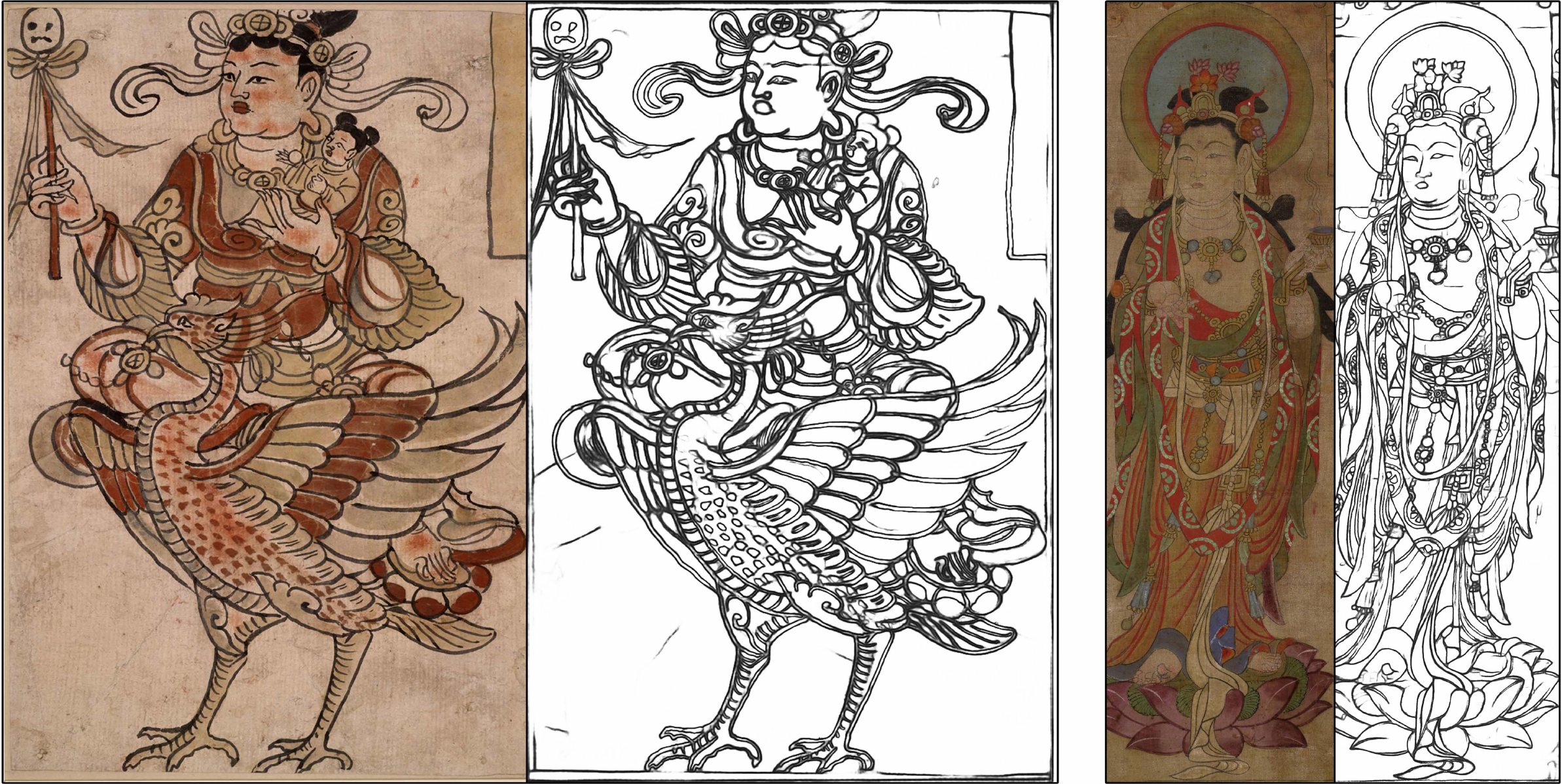}   %./figure/framework.eps
	\caption{$\textbf {Edge Detection Example}$. Our method extracts clear boundaries and edges of Dunhuang murals by exploiting self-attention and convolutional features.}\label{example}
\end{figure}

\begin{figure*}[t]%
	\centering
	\includegraphics[width=0.90\textwidth]{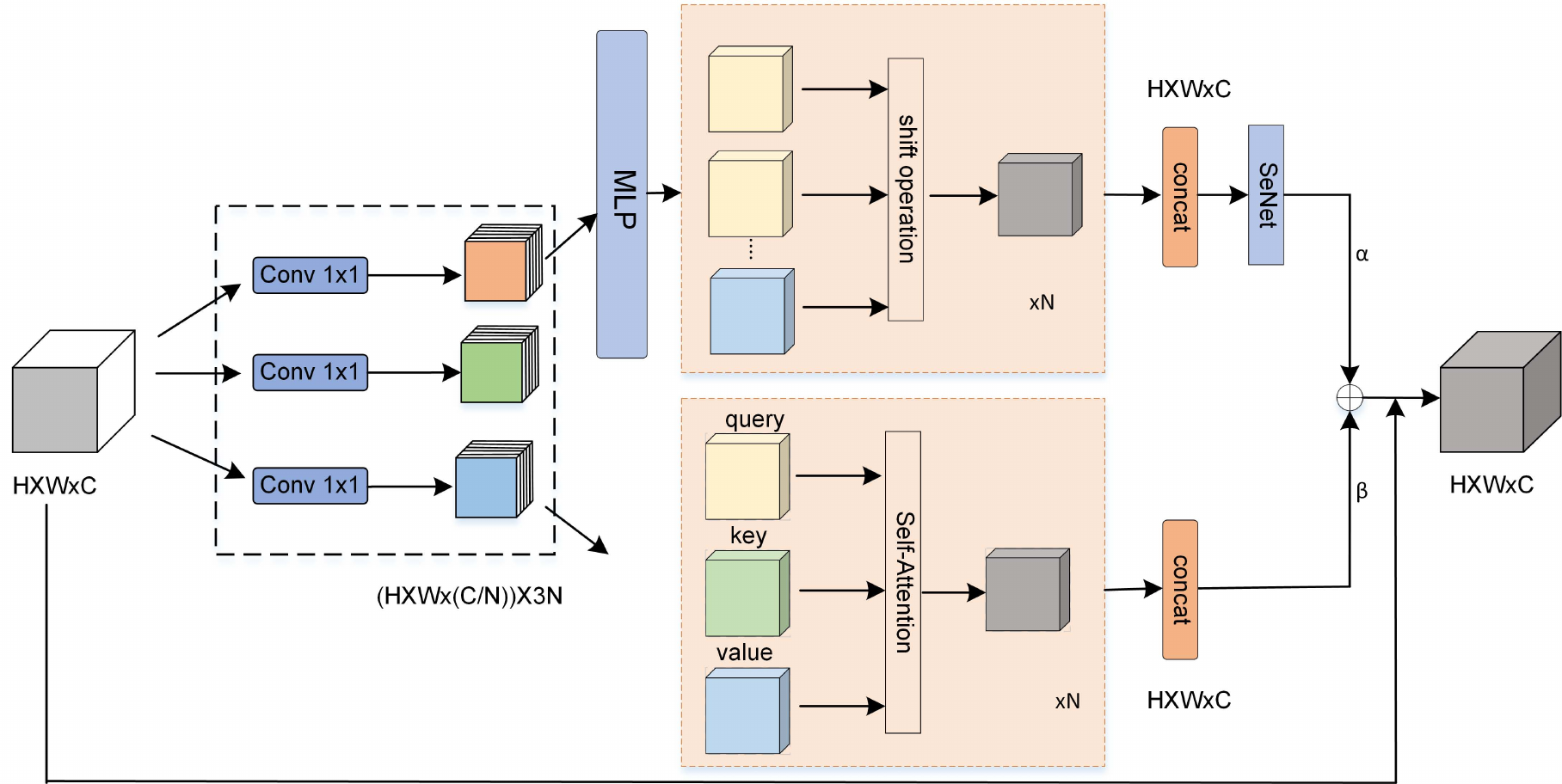}   %./figure/framework.eps
	\caption{An illustration of the proposed Racmix module: We first complete the task of the first stage by projecting the features using three 1*1 convolutions. Since the traditional 3*3 convolution can be decomposed into the sum of 1*1 convolutions that perform different kernel weights, self-attention can get the query, key, and value matrices through 1*1 convolution projection, and the aggregation is obtained by calculation value. Therefore, the intermediate features of convolution and self-attention can be fused, and finally the fusion result is connected with the input feature map using residuals.}\label{figure2}
\end{figure*}

To summarize, this paper makes the following main contributions:
\begin{enumerate}		
	\item We propose a novel edge detector based on the fusion of self-attention and convolution to generate contour maps of Dunhuang murals. As far as we know, this is the first edge detection network based on self-attention combined with convolutions.
	\item The Racmix module, on the one hand, reveals the potential connection between the convolution and self-attention, and on the other hand enables the entire network to take full advantage of the benefits of both convolution and self-attention paradigms.

    \item The cross-layer fusion structure more effectively enables the multi-level fusion of shallow features and deep features, which enriches the feature information in the prediction feature map.
        	
	\item Extensive experiments on four publicly available edge detection datasets demonstrate that our method achieves state-of-the-art results.

%	We extensively evaluate our proposed method on seven benchmark datasets, which validates that our method outperforms state-of-the-art methods in different applications.	
\end{enumerate}

%The rest of this paper is organized as follows. Section \ref{sec:related} reviews the related work. Section \ref{sec:pre} presents the proposed ANLTSC model, including the optimization process, convergence and computational complexity analysis. The analysis of the experimental results is showed in Section \ref{sec:Experiments}. Section \ref{sec:conclusion} concludes the whole paper.
\section{Related work}
\label{sec:related}
As an important branch of computer vision, edge detection has received extensive attention and research in recent years.
Traditional edge detection methods generally use the gradient of the image to obtain the edge of the image, such as Canny, Sobel and Robert operators \cite{canny1986computational,kanopoulos1988design,chaple2015comparisions}. These methods utilize the low-level semantic features of images and are widely used in computer vision \cite{lin2022bio,al2022low}. In addition, some learning-based methods use these low-level features to obtain image edges and contours by training classifiers. Although these methods have achieved good results, they all use low-level features of images and these features are manually made. They are faced with some difficulties in feature selection and the dilemma of lacking some edge semantic information, so the experimental results obtained are not ideal.

In recent years, the convolutional neural network has become more and more popular in computer vision because of their powerful feature extraction ability. Many methods based on convolutional neural networks have become the main edge extraction methods. These methods use depth supervision methods to learn edge features from different levels of image features and fuse these features to generate image contours \cite{liang2021coarse,deng2018learning,he2019bi,hu2019dynamic,huang2017densely,konishi2003statistical}.
Specifically, the HED network \cite{xie2015holistically} obtains rich hierarchical feature information by supervising the side outputs of different scale feature maps, thereby improving the effect of edge detection.
RCF \cite{liu2017richer} fuses the outputs of all convolutional layers into a single overall network framework, resulting in more efficient edge detection results. Inspired by HED and Xception networks, Xavier Soria et al. \cite{poma2020dense} proposed a dexined network that generates a more reasonable thin edge map by embedding the Xception module in the HED network. PiDiNet \cite{su2021pixel} combines traditional edge detection methods using gradients with convolutional neural networks to achieve a method for quickly acquiring the edges of images.
Although convolutional neural networks have strong semantic extraction ability. However, due to the limited receptive field, the global context information is ignored, which makes the generated edge map have a certain noise edge \cite{iandola2014densenet,deng2020deep,shen2015deepcontour,yang2015boundary}.

$\textbf{Vision transformer}$.
The transformer model abandons the traditional CNN and RNN, and the entire network is completely composed of the attention mechanism, which brings a very large performance improvement in natural language processing.
Later, transformers were applied to computer vision tasks, such as object detection, semantic segmentation, image classification, etc. In 2020, the Google team proposed the Vision Transformer(ViT),
which combines computer vision with natural language tasks by chunking pictures, flattening them into sequences, and encoding them \cite{dosovitskiy2020image}. Compared with all algorithms, it achieves the state-of-the-art on image classification tasks.

Our work is inspired by ViT and previous research on edge detection algorithms \cite{yu2018simultaneous,fu2019dual,acuna2019devil}, but there are two main differences compared to the above methods. First of all, as far as we know, the network based on the combination of self-attention and convolution proposed in this paper is the first to be applied to the edge detection task. Secondly, we perfectly fuse the dense network with the Racmix module to generate clearer edge maps by learning local and global context information.

%Our work is inspired by previous work to learn image edge cues on an affordable computationally expensive one-stage framework by combining the powerful semantic extraction capability of convolutions with the long-dependency capture capability of transformers to obtain image edge map.

\section{Proposed method}
\label{sec:pre}

\begin{figure*}[t]%
	\centering
	\includegraphics[width=0.90\textwidth]{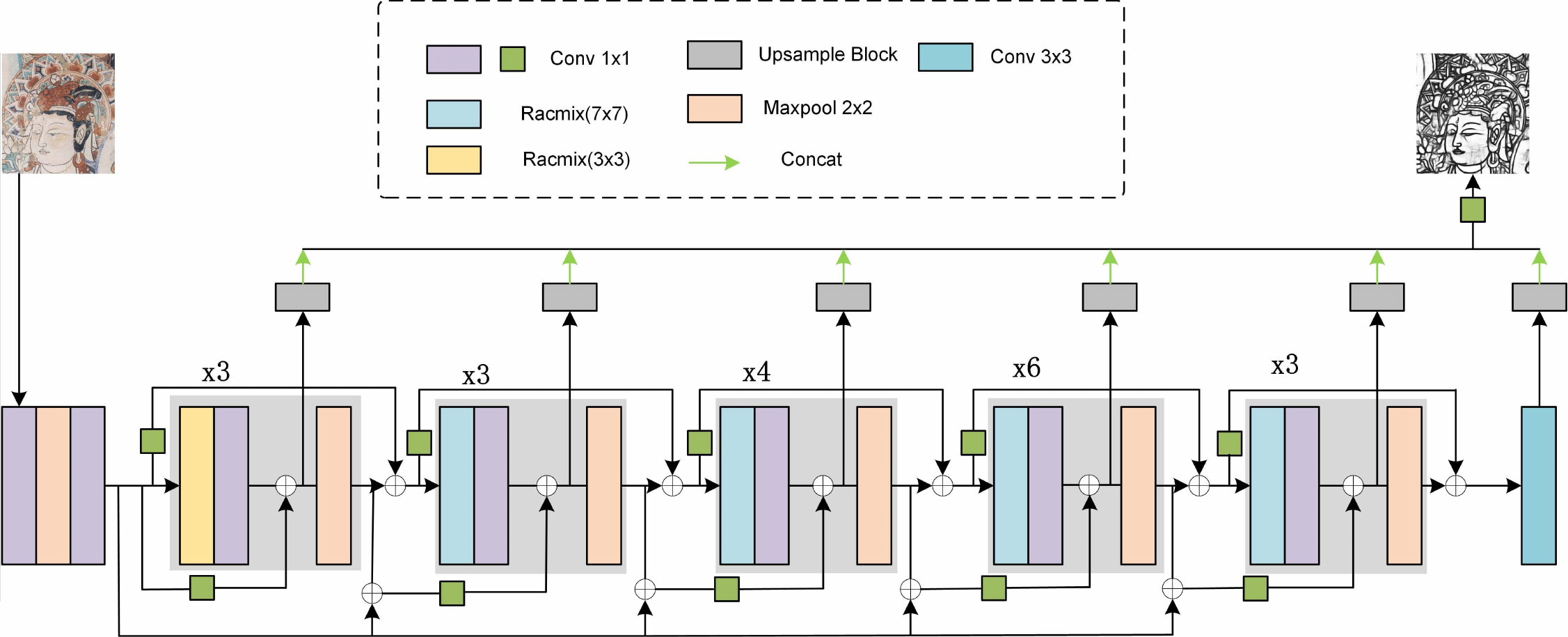}   %./figure/framework.eps
	\caption{Proposed architecture: The network mainly consists of five main modules encoder. The main blocks are connected by residual connections. Each main block consists of a different number of sub-blocks consisting of a Racmix and a 1*1 convolution module. The features of each main block are fed to the next-level main block through residual connections to generate edge maps of different levels, and then the features of different levels are output through the upsampling block, and finally the edge maps of different levels are fused to obtain the final result.}\label{fig:network}
\end{figure*}

In this section, we first propose an edge detection network based on the fusion of self-attention and convolution, which consists of a set of hierarchically stacked blocks that receive feature maps of different sizes,
and then output edge map with the same resolution. The network can be viewed as two distinct parts(see Figure $\ref{fig:network}$ and $\ref{figure2}$): Dense network and  residual self-attention and convolution mixed module (Racmix).
The dense network is mainly used to fuse the shallow features with the deep features to avoid the loss of some shallow information in the deep network. The Racmix module mainly mines the relationship between self-attention and convolution to obtain rich intermediate features. The network does not require a pre-trained model, and in most cases, it achieves state-of-the-art results on the Dunhuang mural dataset, as shown in Figure $\ref{example}$.

%In this section, we first propose an edge detection method based on the fusion of self-attention and convolution. Among them, self-attention can adaptively focus on different regions by calculating the attention weight between related pixels, thereby capturing more feature information, while convolution operation can provide important inductive biases for feature maps by sharing convolution kernels. In this model, the mixing module Racmix is used to fully explore the relationship between the two, and obtain rich intermediate features.

The overall framework of the entire network is shown in Figure $\ref{fig:network}$. It consists of an encoder with five main blocks consisting of Racmix modules and 1*1 convolution stacking blocks. The Racmix module will be introduced in detail in the following chapters.
In order to fuse the features of the shallow layers with the features of the deep layers in the throughout network.
Local and global residual connections are used in this network. The local residual connection is mainly responsible for the direct feature propagation of stacked Racmix modules. Each global residual connection uses dense connections to propagate shallow features to each different stack. In addition, channel information is exchanged between different residual blocks by using $1*1$ convolution.
All edge maps generated by the upsampling block will be spliced together by concat, and then fed to the end of the network to learn and generate a fusion edge map. All six upsampling blocks do not share weights.

\textbf{\subsection{Convolution}}

Convolution is an important part of the convolutional neural network \cite{xu2017learning,zhang2016semicontour,yu2017casenet}. The convolution operation is generally a process in which the convolution kernel slides in the input image to obtain the feature map, in which the convolution kernel performs a convolution operation every time it slides until the final feature map is obtained. In this paper, we regard convolution as performing $k*k$ projections on the input feature map through a convolution kernel of size $k*k$, obtaining projected feature maps at different positions of the convolution kernel, and add these projected feature maps.
Therefore, we can decompose the standard convolution into two steps. In the first stage, the feature map is linearly projected with weights from a certain position, which is generally implemented by $1*1$ convolution. In the second stage, the projected feature map is shifted according to the position of the kernel, and finally these projections are clustered together. In the entire calculation process, except for the $1*1$ convolution operation, the rest are lightweight operations.
The specific calculation method is as follows:

First, for convenience of representation, the stride of the convolution is set to $1$. The convolution kernel is defined as $K \in R^{C_{in}\times C_{out}\times k \times k}$,
where $k$ is the size of the convolution kernel and $C_{in}$, $C_{out}$ are the input and output channel sizes, respectively.
We define $I_{i,j}$, $O_{i,j}$ as the input feature and output feature. A standard convolution can be defined as:
\begin{align}\label{eq:cnn}
O_{i j}=\sum_{p, q} K_{p, q} I_{i+p-\lfloor k / 2\rfloor, j+q-\lfloor k / 2\rfloor}
\end{align}
where $K_{p, q} \in \mathcal{R}^{C_{\mathrm{out}} \times C_{\mathrm{in}}}$ represents the weight on the feature map position $(p,q)$.
We define the shift operation as follows:
\begin{align}\label{eq:cnn1}
\tilde{I}_{i, j}=I_{i+\Delta x, j+\Delta y}, \forall i, j
\end{align}
where $\Delta x$, $\Delta y$ is the displacement distance in the horizontal and vertical directions.

Therefore, the standard convolution can be summarized as:
\begin{align}\label{eq:cnn3}
\tilde{O}_{i j}^{(p, q)}=K_{p, q} I_{i j}
\end{align}
where $I_{i,j}$ represents the input feature map, $K_{p, q}$ is the weight on the feature map position $(p,q)$, $\tilde{O}_{i j}^{(p, q)}$ is the weighted output feature map.
\begin{align}\label{eq:cnn4}
O_{i j}^{(p, q)}=\operatorname{Shift}\left(\bar{O}_{i j}^{(p, q)}, p-\lfloor k / 2\rfloor,q-\lfloor k / 2\rfloor\right)
\end{align}
where $Shift(\cdot)$ indicates a shift operation, $O_{i j}^{(p, q)}$ is all feature maps output after the shift operation.
\begin{align}\label{eq:cnn5}
O_{i j}=\sum_{p, q} O_{i j}^{(p, q)}
\end{align}
where $\sum$ represents the concatenation operation of the channel dimension for all the shifted output feature maps.

In the first stage, the input feature map is a linear projection. The kernel weight is multiplied by the feature tensor of the input pixel to obtain the weighted feature vector. In the second phase, the projected feature map moves according to the kernel location and aggregates all the results together.
The convolution can be decomposed into a linear projection of the weights through Eq $(\ref{eq:cnn3})$, Eq $(\ref{eq:cnn4})$, Eq $(\ref{eq:cnn5})$, the kernel positions are shifted, and finally aggregated together.

In order to realize the shift operation in convolution, we use $k*k$ fixed-size weight matrices for convolution calculation to realize the shift operation of tensors. For the fusion of different directional features, we
all input feature maps are concatenated with convolution kernels, and the entire shift operation becomes a single group convolution.

\textbf{\subsection{Self-attention}}
Attention mechanisms are extensively used in vision tasks. Compared with traditional convolution, attention can obtain larger receptive field information \cite{hu2019dynamic}. In the vision transformer paper, the image is first divided into $N*N$ image sequences, and then converted into a one-dimensional $H/N*W/N$ visual token. These sequences are mapped to the latent embedding space through fully connected layers. To preserve location information, a learnable 1D location embedding is added to the patch. Finally, these features are fed into the transformer encoder. Each standard transformer encoder consists of a multi-head self-attention and a fully connected layer. The self-attention obtains three vectors $Q$, $K$ and $V$ through the embedding vector and calculates the corresponding weight value through the formula \cite{chen2022mixformer,zheng2021rethinking,zhu2020deformable,zhao2020exploring}.

\begin{align}\label{eq:qkv}
Q=W^qI,K=W^kI,V=W^vI
\end{align}
where $W^k$, $W^q$, $W^v$ are learnable parameter matrices.

\begin{align}\label{eq:rtlrr1}
y=softmax(\frac{QK^T}{\sqrt{d_k}})V
\end{align}
where $y$ is the computed attention weight, Q is the query vector, K is the check vector, V is the content vector.

The above mainly describes the calculation process of the single-head self-attention mechanism. In this paper, in order to realize the fusion of convolution and self-attention module, the calculation of self-attention mechanism is described as the following process:

Firstly, the input feature map and output feature map are defined as~$F \in \mathcal{R}^{C_{\text {in }} \times H \times W}$ and $G \in \mathcal{R}^{C_{\text {out }} \times H \times W}$,
Secondly, $f_{i j} \in \mathcal{R}^{C_{\text {in }}}$ and $g_{i j} \in \mathcal{R}^{C_{\text {out }}}$
denote the input and output feature maps of the corresponding pixel $(i, j)$. Finally, output of the
attention module is computed as:
\begin{align}\label{eq:rtlrr2}
g_{i j}=MLP \left( \prod_{l=1}^N\left(\sum_{a, b \in \mathcal{N}_k(i, j)}
 \mathrm{A}\left(W_q^{(l)} f_{i j}, W_k^{(l)} f_{a b}\right) W_v^{(l)} f_{a b}\right) \right)
\end{align}
where $\prod$ donates the concat operation on the output of the $N$ attention heads. $MLP$ stands for linear transformation operation.
$W_q^{(l)}, W_k^{(l)}, W_v^{(l)}$ donate queries, keys, values of the projection matrix, respectively. $\mathcal{N}_k(i, j)$
represents the pixel local area with $(i, j)$ as the center and $k$ as the radius and $\mathrm{A}\left(W_q^{(l)} f_{i j}, W_k^{(l)} f_{a b}\right)$ is the corresponding attention weight of features within $\mathcal{N}_k(i,j)$.
\begin{align}\label{eq:rtlrr3}
\mathrm{A}\left(W_{q}^{(l)} f_{i j}, W_{k}^{(l)} f_{a b}\right) =
\operatorname{softmax}_{\mathcal{N}_{k}(i, j)}\left(\frac{\left(W_{q}^{(l)} f_{i j}\right)^{\mathrm{T}}\left(W_{k}^{(l)} f_{a b}\right)}{\sqrt{d}}\right)
\end{align}
where $d$ is the feature dimension.

The operation of self-attention is similar to convolution. First, $1*1$ convolution is performed, and the input feature map is projected as query, key, and value. Then calculate the sum of the attention weights and the aggregated value matrix.

It can be seen from the above observations that both the convolution and self-attention modules perform a $1*1$ convolution operation. Therefore, we can share $1*1$ convolution operations, perform only one projection operation, and perform different aggregation operations on these feature maps.

Specifically, the Racmix module contains two stages. In the first stage, we use three $1*1$ convolutions for projection and reshape them into $N$ feature maps respectively. In this way, we will obtain intermediate results of $3*N$ feature maps. In the second stage, two different paths use different ways to aggregate operations. For the convolutional path, we can generate $k^2$ feature maps through fully connected layers. Features are then generated by moving and merging operations. For the self-attention path, we divide the intermediate results into $N$ groups, and each group contains three features. These three features are used as query, key, value matrices respectively, and the calculation formula of the multi-head self-attention module is used to obtain the final weight result.

Finally, we merge the features on the two paths, and the formula is as follows:

\begin{align}\label{eq:rtlrr3}
F_{out}=\sigma(\alpha F_{att}(x)+\beta F_{conv}(x)+x)
\end{align}
where the value of $\alpha$ and $\beta$ are two learnable weight parameters, $\sigma$ is the Relu activation function.
\textbf{\subsection{Shift operation}}
The convolution path is realized by shifting and summing operations. If we move the input tensor features in different directions, the local features of the tensor data will be destroyed and the reasoning speed of the model will be slowed down. To solve this problem, we use deep convolution of fixed kernel to implement shift operation of tensor. The convolution kernel (k=3) is represented as following:
\begin{align}\label{eq:r3}
K_c=\begin{bmatrix}
  1& 0 & 0\\
  0& 0 & 0   \\
  0& 0 & 0
\end{bmatrix}
\end{align}
Then the corresponding output can be expressed as:
\begin{align}\label{eq:r4}
f_{c,p,q}=\sum_{p,q\in\{0,1,2\}}K_{c,p,q}I_{c,i+p-\left \lfloor k/2\right \rfloor,j+q-\left \lfloor k/2 \right \rfloor}
\end{align}

A convolution operation by using this particular convolution kernel is equivalent to a shift operation of the tensor. In order to improve the efficiency of parallel computing, we concatenate all input features and convolution kernel together, and adopt a grouping convolution method to combine the sum of features in different directions.

\textbf{\subsection{Loss Function}}
We use a balanced cross-entropy loss function for deep supervision on each edge map.
The loss is calculated as:
\begin{align}\label{eq:loss}
l(X_{i,j};W)=-\sum_{i,j}^{I}[\alpha G(i,j)\log P(i,j)+ \beta(1-G(i,j))\log(1-P(i,j))]
\end{align}

where $G(i,j)\in \{0,1\}$ is the groud truth label of the pixel$(i,j)$ and $P(i,j)$ is the predicted
probability of edge.
$Y^+$ and $Y^-$ represent the number of positive and negative samples,
then $\alpha= \frac{|Y^-|}{|Y^+|+|Y^-|}$,
$\beta=\frac{\left|Y^{-}\right|}{|Y^+|+\left|Y^{-}\right|}$.

The total loss of the entire network is as follows:
\begin{align}\label{eq:loss2}
L(W)=\sum_{k=1}^{K}l(X_{i,j}^{(k)};W)+l(X_{i,j}^{fuse};W)
\end{align}
where $K$ represents the number of side outputs of the network, which is six. $X_{i,j}^{(k)}$ represents the activation value of the output feature map of the k stage
while $X_{i,j}^{fuse}$ represents the output after feature fusion of different stages.

\section{Experiments and analysis of results}
\label{sec:Experiments}

In order to evaluate the performance of the edge detection algorithm proposed in this paper, four datasets were selected.
They are BSDS500, BIPED, NYUD v2 and Multicue. We will quantitatively evaluate the three commonly used evaluation indicators of $ODS$, $OIS$ and $AP$, and visualize the realization results on different datasets.

In addition, in order to verify the effect of this method in the complex contour image, we cut out the clear images of the Buddha figures in the Dunhuang mural dataset. Compare with several common algorithms, the experimental results show that our method can extract the outline of the clearer murals.

\textbf{\subsection{Implementation Details}}
We use pytorch to implement the model. Firstly, we randomly combine the BIPED dataset by using crop, scale, rotate and
 gamma corrections operations. The specific process is as follows: i) Divide the original high-resolution image into two parts according to the width. ii) Each image is rotated at 15 different angles and cropped to a 700*700 resolution image according to the center of the image. iii) Each image is flipped horizontally and vertically. iv) Apply two gamma corrections ($0.3030$, $0.6060$). The original $200$ BIPED dataset was expanded to $21600$ as the training dataset.
Secondly, we train the entire model by using the Adam optimizer with learning rate 1e-6, weight decay 1e-8, and batchsize $8$. It takes about $11$ days to perform $800$ epochs on a 2080Ti GPU, and the model basically tends to converge.

\textbf{\subsection{Datasets description}}

$\textbf{BSDS500}$ is a dataset provided by the University of Berkeley, which is mainly used for image segmentation and edge detection tasks. The dataset contains $500$ natural images of
481*321 pixels each, including  $200$ training images, $100$ validation images, and $200$ testing images, each of which has been manually annotated by 5 people.

$\textbf{BIPED}$ is annotated by experts in the field of computer vision and mainly consists of $250$ outdoor photos, each of which is $1280*720$ pixels, of which $200$ are used for training and $50$ are used for testing.

$\textbf{NYUD v2}$ is a set of data sets provided by New York University. The dataset has a total of $1449$ images and contains $464$ indoor scenes. Mainly used in the field of scene segmentation. In this paper, $500$ images are randomly selected as the test set for evaluation.

$\textbf{Multicue}$ is composed of $100$ outdoor high-definition pictures of size $1280*720$, in which each scene has $5$ boundary annotations and $6$ edge annotations. Among them, $80$ are training sets and $20$ are test sets. In this paper, $20$ images were randomly selected as the test set for evaluation.

$\textbf{Dunhuang murals}$. This dataset is mainly based on Dunhuang murals in the prosperous Tang Dynasty, and the original image is cut to a size of 512*512. We selected relatively clear and complete murals, mostly figures of Buddha statues, and got a total of 3,000 murals.

\textbf{\subsection{Evaluation metrics}}
In order to evaluate the performance of the edge detection algorithm, we adopted several commonly used evaluation indicators, including $ODS$ (Optimal Dataset Scale), $OIS$ (Optimal Image Scale), $AP$(Average Precision).

$\textbf{ODS}$ stands for Global Optimal Threshold, and a fixed threshold is used for all images in the dataset. That is, a threshold is selected to maximize the $F$ value obtained on all images. The $F$ value is defined as the harmonic mean of Precision and Recall, which is expressed by the formula as :
\begin{align}\label{eq:f1}
F=(1+\beta^2)\cdot \frac{Precision\cdot Recall}{\beta^2\cdot Precision+Recall}
\end{align}

$\textbf{OIS}$ represents the optimal threshold for each image, that is, selecting an optimal threshold for each image to maximize the $F$ value of the image, and its $F$ value is defined as above.

$\textbf{AP}$ stands for Average Accuracy, which refers to the integral of the area under the $PR$ curve. In the process of code implementation, it is generally obtained by sampling and averaging on the $PR$ curve.

In edge detection, it is necessary to use a concept called distance tolerance to determine whether the pixels of the predicted boundary are correctly predicted. The so-called distance tolerance refers to allowing a small positioning error between the real boundary and the predicted boundary. Generally, a maxdist variable is used in the code to represent, and the default setting is 0.0075. This variable is multiplied by the width and height of the image to get the distance tolerance of the image in the column and row directions.

\textbf{\subsection{Performance evaluation}}
We quantitatively compare our approach with several common convolution-based edge detection methods, including the well-known HED \cite{xie2015holistically}, RCF \cite{liu2017richer}, and BDCN \cite{he2019bi}. Among them, HED is based on vgg-16 networks, and the edge detection is fixed to integrate the side output features of 5 stages to achieve edge detection. On the basis of HED, the RCF network is combined with all convolutional layer information together to obtain richer features in different standards, which greatly improves the performance of image edge detection.
BDCN is a bi-directional network. It uses the edge of specific scale to monitor each layer output. In order to enhance the characteristics of each layer output, the SEM module is used to generate multi-scale features, and the multi-scale features are finally fused.
Inspired by HED and Xception, a new edge model DexiNed is proposed. The network consists of an encoder and an upper sample block, and finally the output is fused to get the final edge map.
EDTER is an edge detector based on transformer. It mainly uses the context information and local clues of the image to extract clear and meaningful edges.
For the fairness of the comparison results, the edge maps generated by all comparison algorithms are subjected to non-maximum suppression operations. The specific comparison results are shown in  Table $\ref{tab:all2}$.

\begin{figure*}[h]% h ?????t?? ;b???p ???; ???[tbp]		
	\centering
	\subfigcapskip=5pt % ?????????????
	\subfigure[Input ]{
		\begin{minipage}{0.12\linewidth}	
			\centering
			\includegraphics[width=2.2cm]{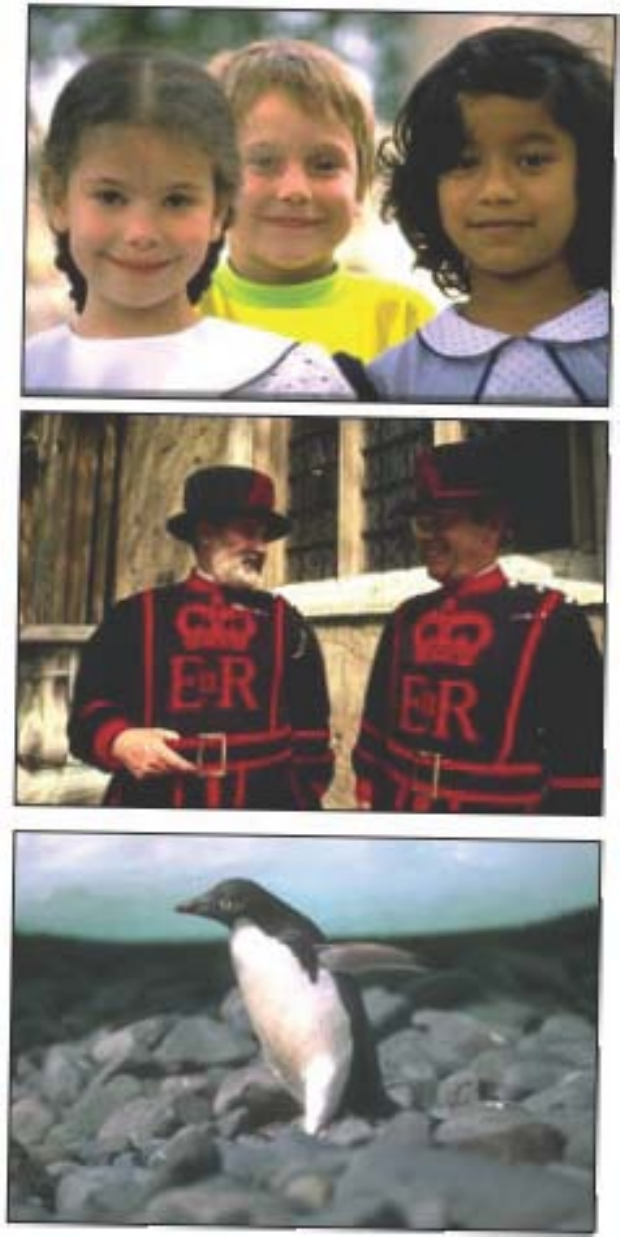}
		\end{minipage}
	}
    	\subfigure[GT ]{
		\begin{minipage}{0.12\linewidth}	
			\centering
			\includegraphics[width=2.2cm]{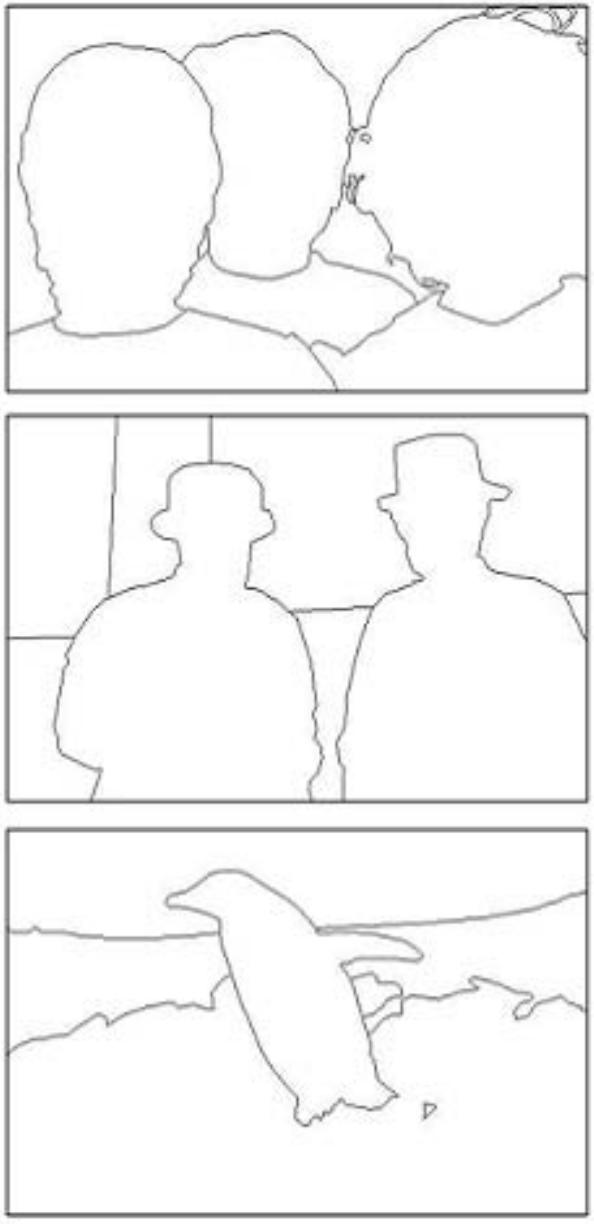}
		\end{minipage}
	}
	\subfigure[RCF ]{
		\begin{minipage}{0.12\linewidth}	
			\centering
			\includegraphics[width=2.2cm]{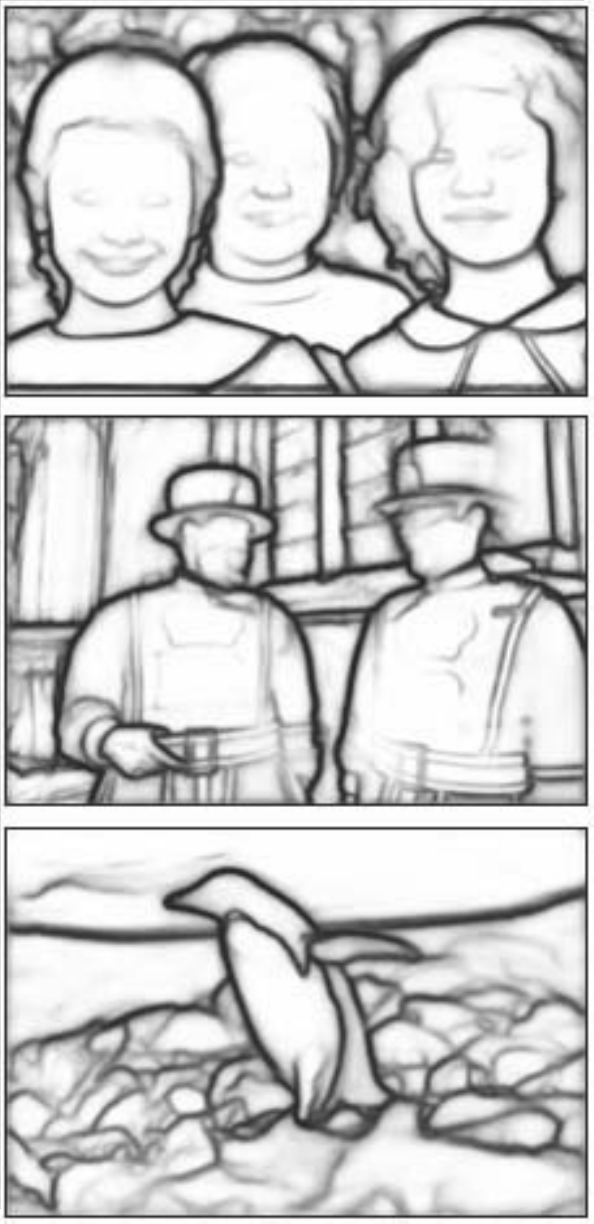}
		\end{minipage}
	}
	\subfigure[BDCN]{
		\begin{minipage}{0.12\linewidth}	
			\centering
			\includegraphics[width=2.2cm]{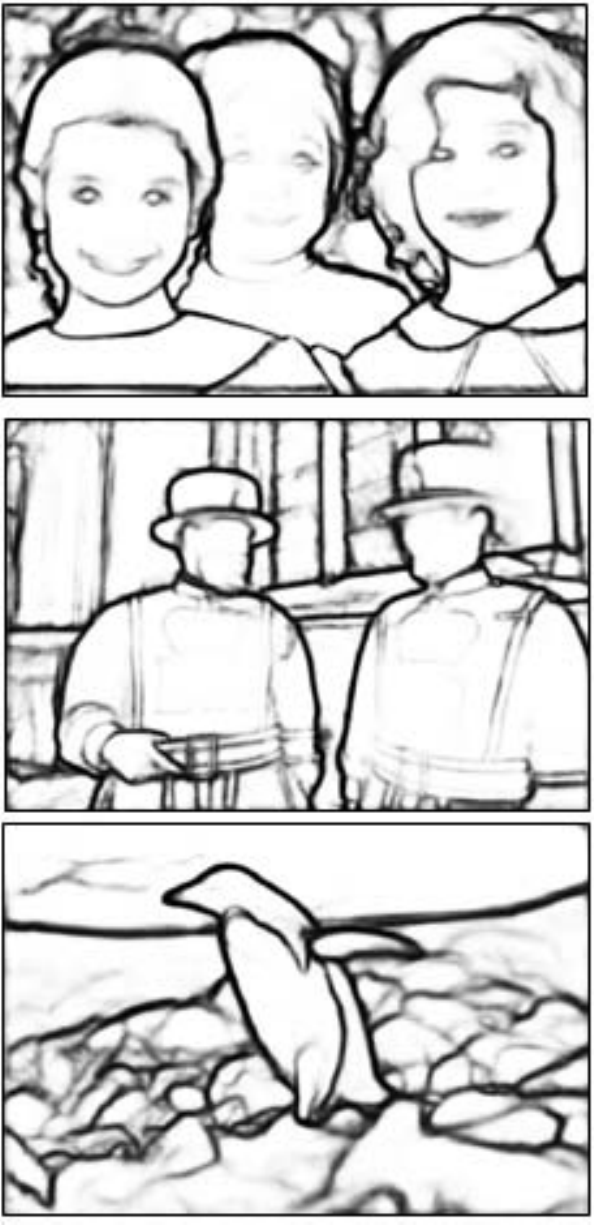}
		\end{minipage}
	}
	\subfigure[EDTER]{
		\begin{minipage}{0.12\linewidth}	
			\centering
			\includegraphics[width=2.2cm]{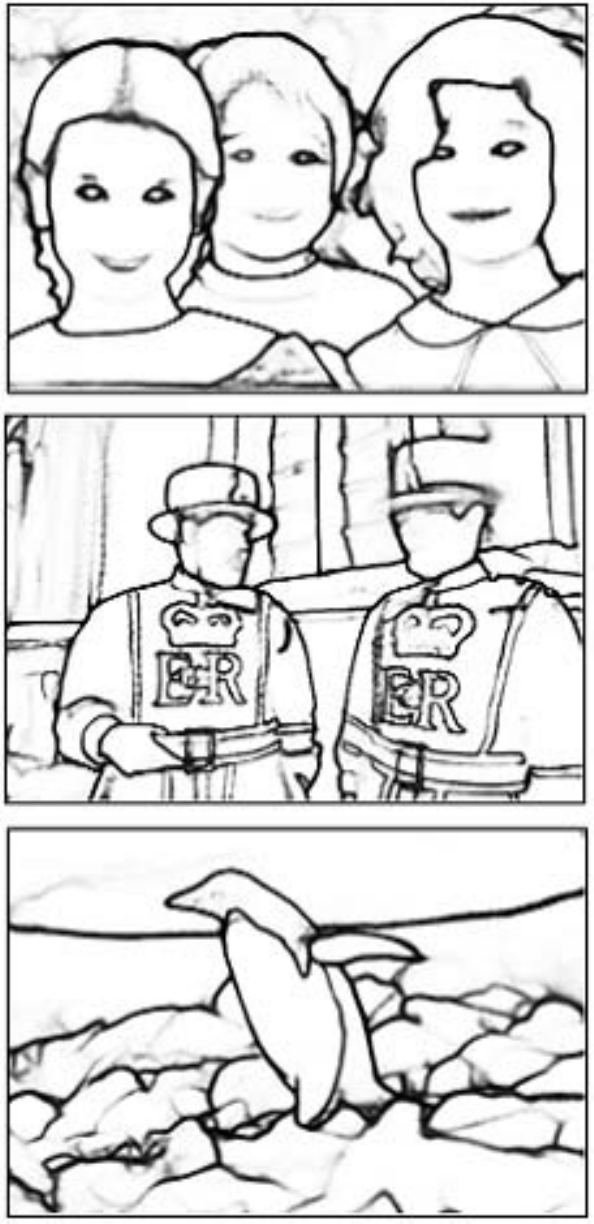}
		\end{minipage}
	}
    	\subfigure[DEXINIED]{
		\begin{minipage}{0.12\linewidth}	
			\centering
			\includegraphics[width=2.2cm]{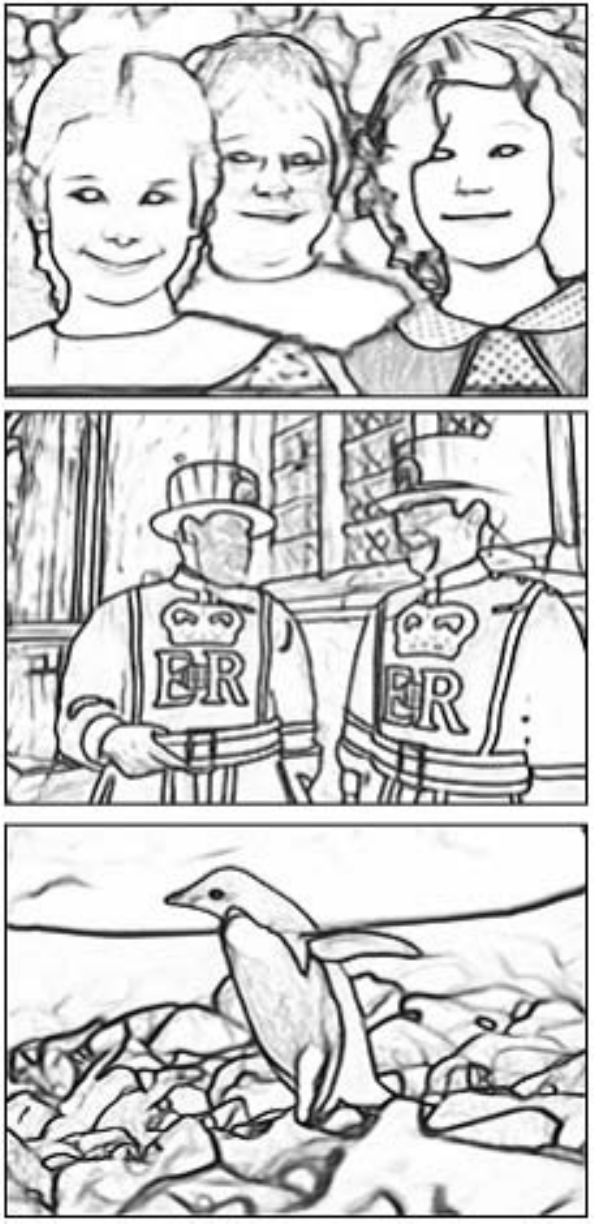}
		\end{minipage}
	}
	\subfigure[OUR ]{
		\begin{minipage}{0.12\linewidth}	
			\centering
			\includegraphics[width=2.2cm]{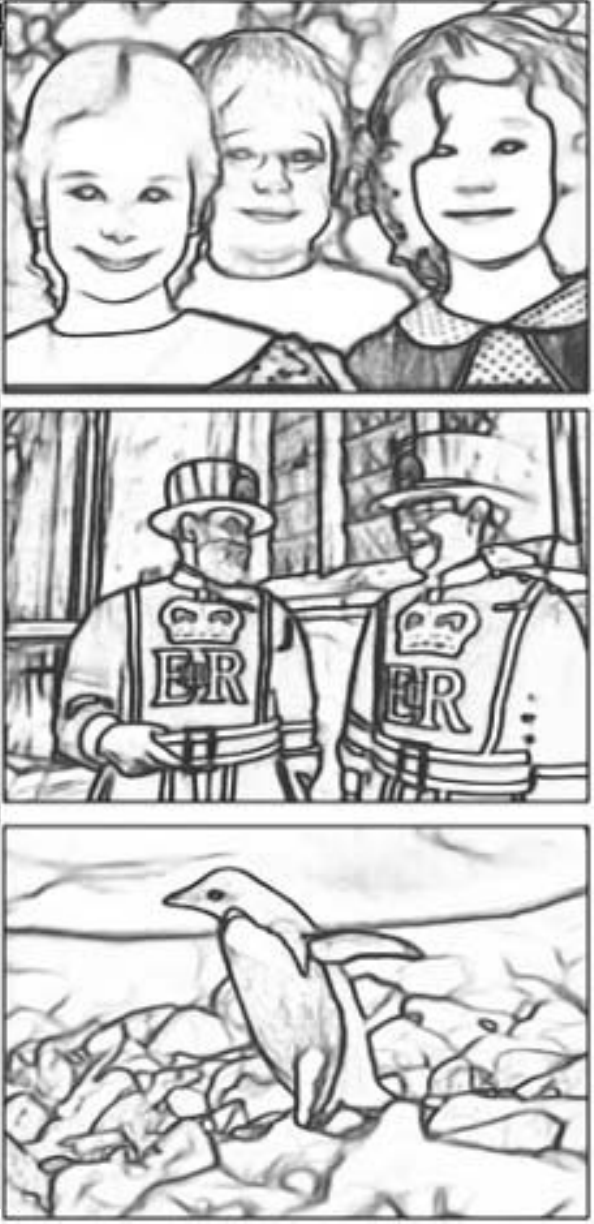}
		\end{minipage}
	}
	\hspace{0.1 cm}
	\caption{Qualitative comparisons on three challenging samples in the testing set of BSDS500.}
	\label{fig:bsds500}
\end{figure*}

$\textbf{On BSDS500}$. We compare with traditional and deep learning based edge detection algorithms on the BSDS500 dataset. The F-measure $ODS$ of our model reaches $78.2\%$, which is clearly lower than the detection results of several other algorithms. The main reason is that GT on the BSDS500 dataset gives relatively few edge lines, while our proposed algorithm detects more edge lines, so the value of our proposed algorithm is low in the evaluation indicator.
In order to more intuitively prove the correctness of our idea, we have visualized the detection results of different algorithms. In Figure $\ref{fig:bsds500}$, our detection results have more details than GT, such as the facial expressions of the characters in the figure, the letters on the clothes, and the outlines of stones in the penguin background.
From the above qualitative and quantitative results, it can be seen that our model has obvious advantages in image detail detection, which can achieve both clear and accurate results.

$\textbf{On NYUD v2}$. In NYUD v2, we conduct experiments on the test dataset and compare with several state-of-the-art edge detection algorithms, such as Canny \cite{canny1986computational}, HED \cite{xie2015holistically}, RCF \cite{liu2017richer}, BDCN \cite{he2019bi}, PiDiNet \cite{su2021pixel}, RINDNet \cite{pu2021rindnet} and EDTER \cite{pu2022edter}. The quantitative results of our method and several other algorithms are shown in Table $\ref{tab:all2}$. Our method achieves the best scores of $78.7\%$, $79.1\%$, and $81.1\%$ on $ODS$, $OIS$, and $AP$. Compared with the second best EDTER algorithm, the three indicators of our algorithm have improved by $1.3\%$, $0.2\%$, and $1.4\%$ respectively.

$\textbf{On Multicue}$. The dataset contains two annotation maps, Multicue boundary and Multicue edge.
In this paper, the Multicue edge is selected as the annotation map for experiments. As shown in Table $\ref{tab:all2}$, our method shows very competitive results compared with several other algorithms. Among them, the F-measure $ODS$ and $OIS$ achieve $ 89.6\%$ and $90.2\%$ respectively, which are better than other edge detection algorithms.

$\textbf{On BIPED}$. This dataset is a very carefully annotated edge dataset.
Each image in the dataset is carefully annotated and cross-checked by experts in the field of computer vision, and can be used as a benchmark dataset for evaluating edge detection algorithms.
We randomly select $50$ of these images from this dataset for testing.
The quantitative results are shown in Table $\ref{tab:all2}$ and Figure $\ref{fig:pr}$ shows the PR curves for all methods.
Compared with several other algorithms, it can be seen that our method achieves the best scores in $ODS$, $OIS$, $AP$ indicators, which are $87.8\%$, $89.3\%$, $91.1\%$, respectively. Experimental results demonstrate that our method outperforms all existing edge detection algorithms. It further proves that the method proposed in this paper has better performance in extracting image detail edges.

$\textbf{On Dunhuang murals}$. In this paper, we have selected some murals of Buddha figures with complex outlines for experimental tests. As can be seen from the visualization results in Figure $\ref{fig:mural}$, our method achieves the best edge maps on Buddha figures, backlights, and headlights.
As a representative of traditional edge detection algorithms, Canny detector produces some false and discontinuous edges due to the lack of certain semantic feature information.
For RCF \cite{liu2017richer}, BDCN \cite{he2019bi}, DexiNed \cite{poma2020dense}, RINDNet \cite{pu2021rindnet}, these deep learning-based edge detection methods also failed to generate some reasonable edges(especially the headgear of Buddha statues).
From the edge detection results of several other deep learning, it can be seen that the lines in the Buddha figure's headgear are all missing to a certain extent in the first row of Figure $\ref{fig:mural}$; In the second row, the contours of the headgear and facial features of the Buddha are not detected in RCF and BDCN; In the third row, the details of the Buddha figure's facial expressions are missing, especially the eye area is more severe.
On the contrary, our method generates almost all important edge features and shows the contouring effect of murals well. In addition, we present the visualization results of the line drawings of local patterns of Dunhuang murals in Figure $\ref{fig:mural2}$. In the first row of Figure $\ref{fig:mural2}$, the costumes, hands, and instruments of the Buddha statue have missing and blurred lines in Canny, RCF, and BDCN methods, among which DexiNed and RindNet perform better online There are different degrees of burrs on the edges of the bars, and the lines are messy, but the method in this paper extracts relatively clear and semantic contours. From the second, third, and fourth lines of murals in Figure $\ref{fig:mural2}$, it can be seen that the facial expressions of Buddha statues, the detailed textures of feathers and feet of mythical animals are lost and blurred in different ways in RCF, BDCN, DexiNed, and RindNet algorithms. However, our algorithm distinguishes and extracts the lines of the details in the murals very well, thus generating more realistic and artistically appreciative line drawings of Dunhuang murals.

\begin{figure*}[h]% h ?????t?? ;b???p ???; ???[tbp]		
	\centering
	\subfigcapskip=5pt % ?????????????
	\subfigure[Input ]{
		\begin{minipage}{0.12\linewidth}	
			\centering
			\includegraphics[width=2.2cm]{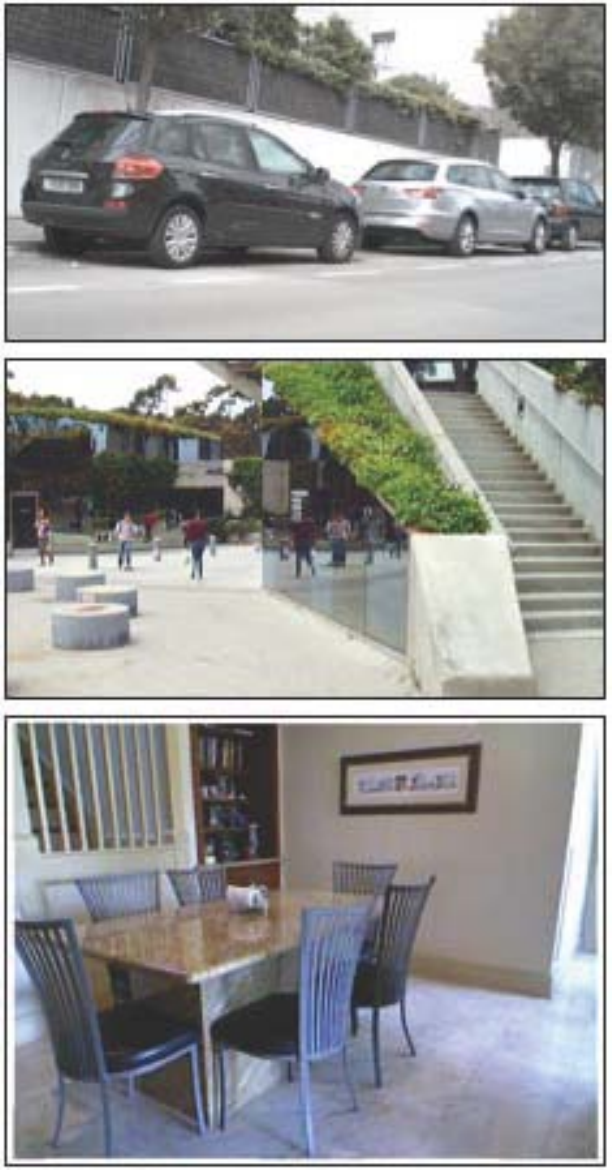}
		\end{minipage}
	}
	\subfigure[GT ]{
		\begin{minipage}{0.12\linewidth}	
			\centering
			\includegraphics[width=2.2cm]{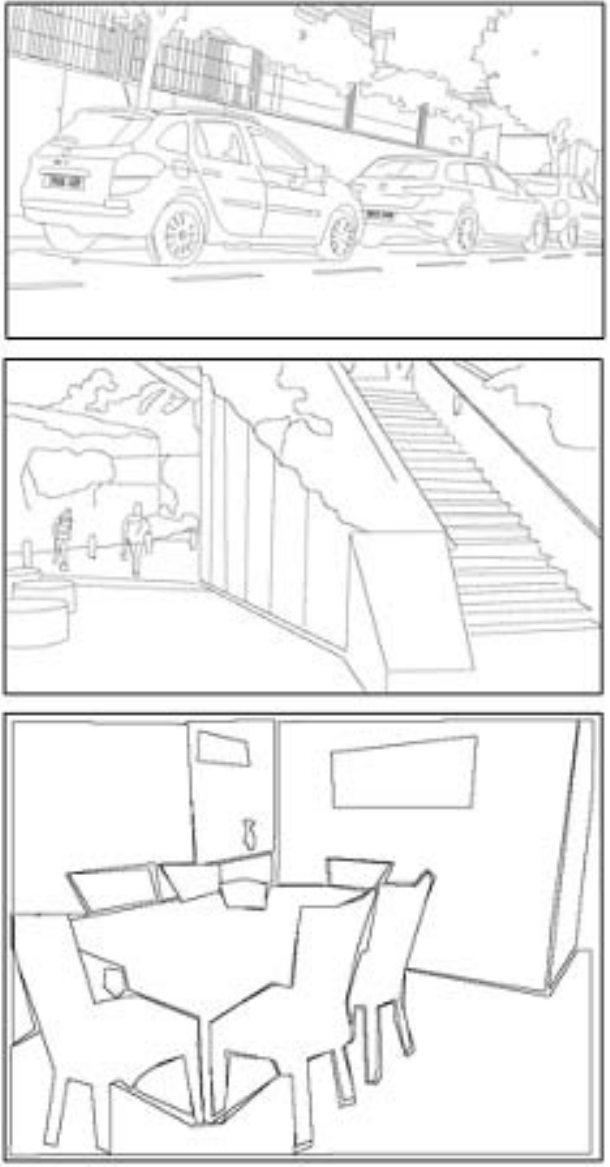}
		\end{minipage}
	}
	\subfigure[RCF ]{
		\begin{minipage}{0.12\linewidth}	
			\centering
			\includegraphics[width=2.2cm]{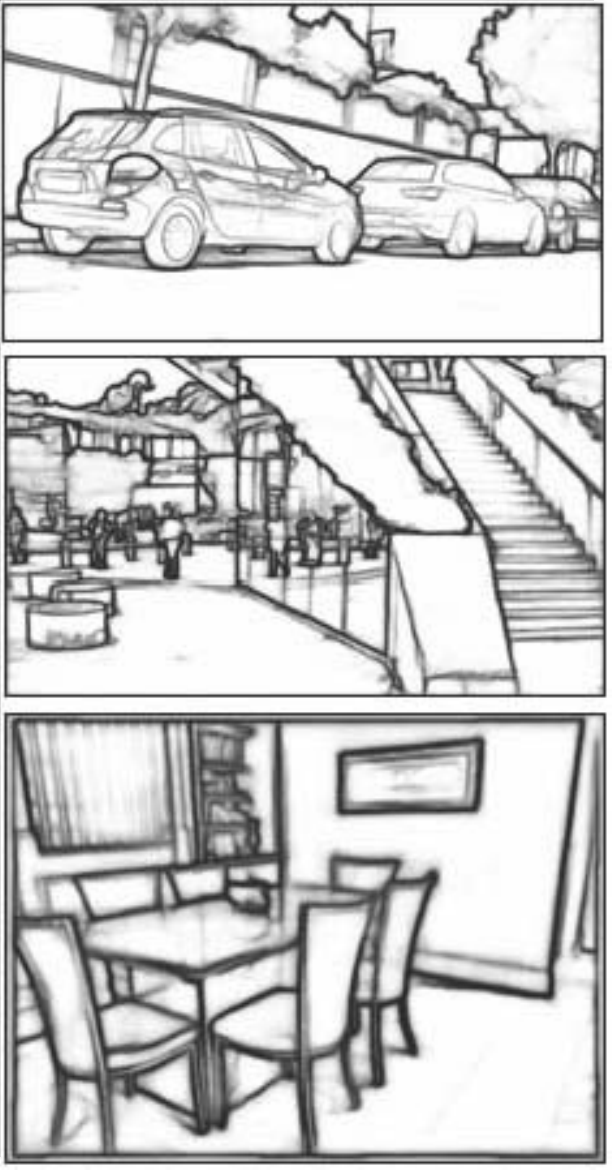}
		\end{minipage}
	}
	\subfigure[BDCN]{
		\begin{minipage}{0.12\linewidth}	
			\centering
			\includegraphics[width=2.2cm]{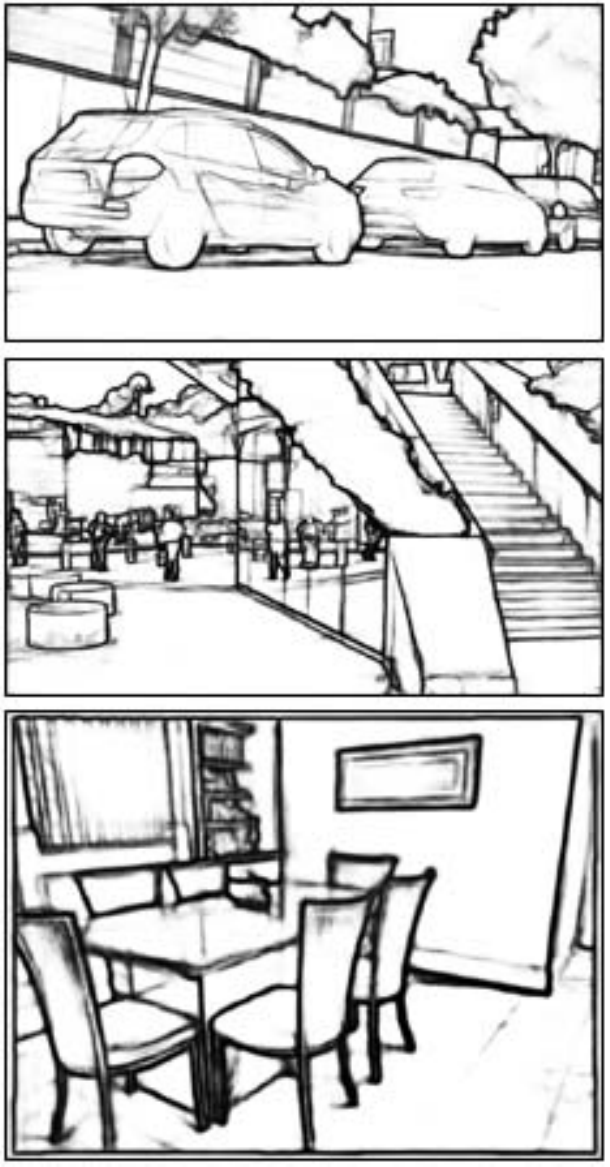}
		\end{minipage}
	}
	\subfigure[RINDNet]{
		\begin{minipage}{0.12\linewidth}	
			\centering
			\includegraphics[width=2.2cm]{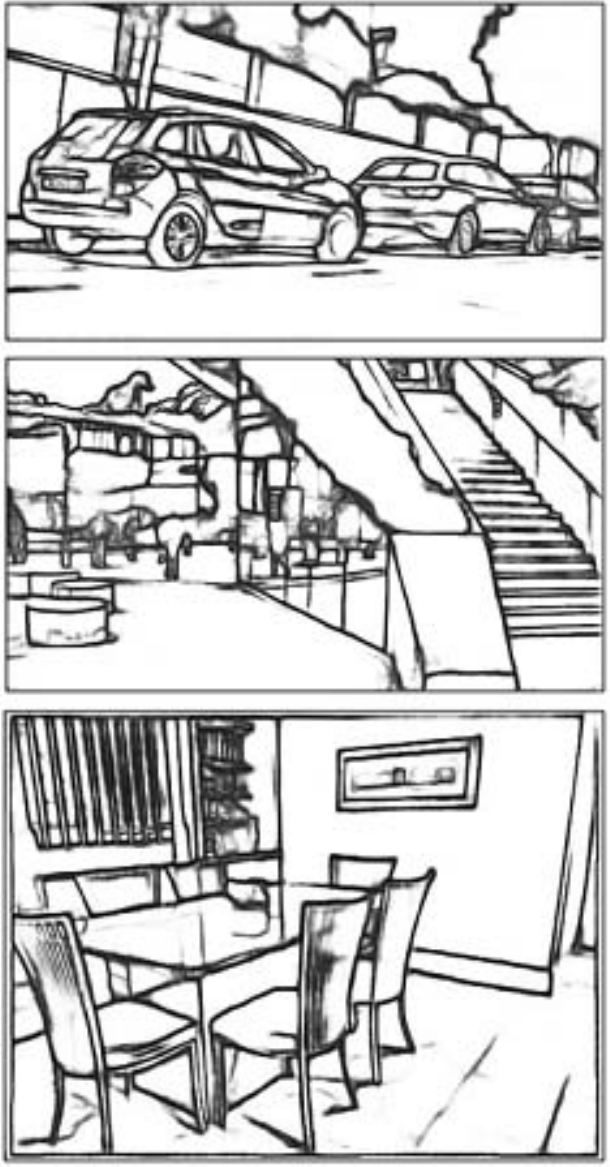}
		\end{minipage}
	}
	\subfigure[DEXINED]{
		\begin{minipage}{0.12\linewidth}	
			\centering
			\includegraphics[width=2.2cm]{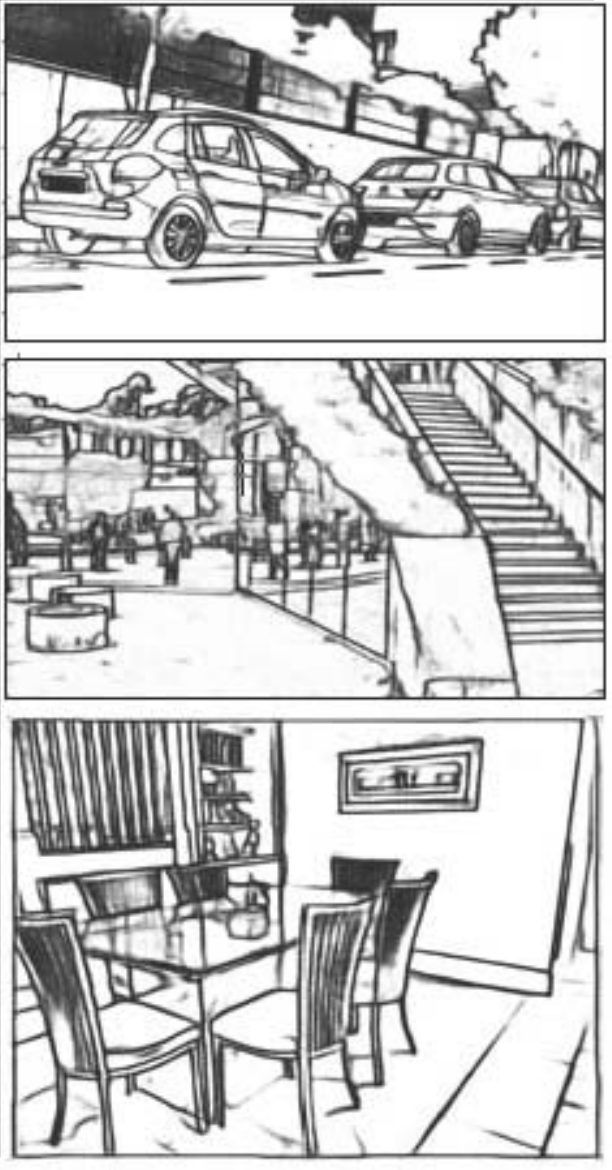}
		\end{minipage}
	}
	\subfigure[OUR ]{
		\begin{minipage}{0.12\linewidth}	
			\centering
			\includegraphics[width=2.2cm]{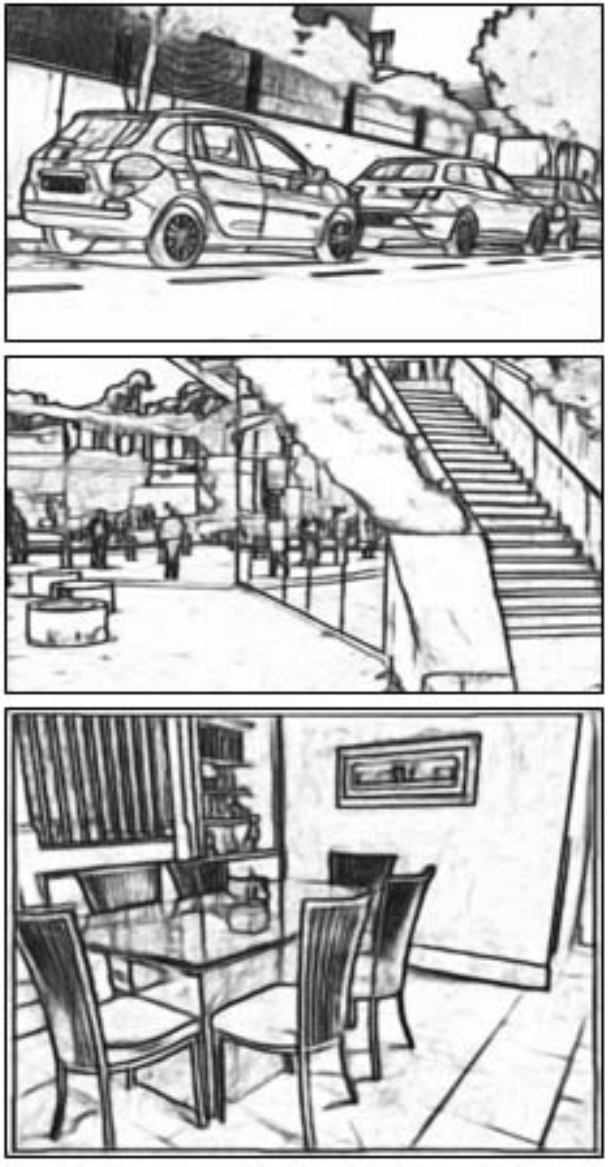}
		\end{minipage}
	}
	\hspace{0.1 cm}
	\caption{The visual effects of several different algorithms on the three datasets are displayed in the figure. From top to bottom are BIPED dataset, Multicue dataset, NYUD v2 dataset.}
	\label{fig:img_data}
\end{figure*}

\begin{figure*}[h]% h ?????t?? ;b???p ???; ???[tbp]		
	\centering
	\subfigcapskip=5pt % ?????????????
	\subfigure[Input ]{
		\begin{minipage}{0.12\linewidth}	
			\centering
			\includegraphics[width=2.2cm]{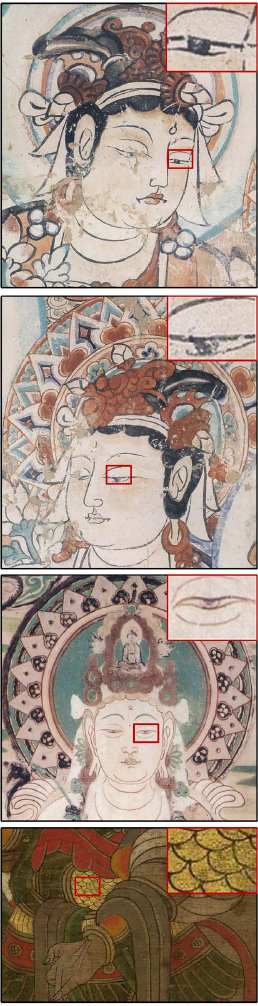}
		\end{minipage}
	}
	\subfigure[Canny]{
		\begin{minipage}{0.12\linewidth}	
			\centering
			\includegraphics[width=2.2cm]{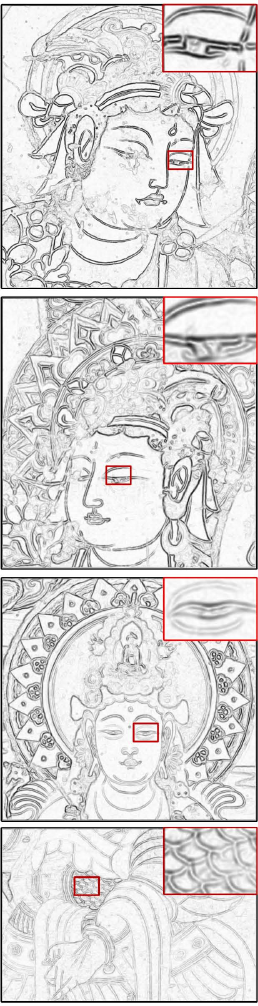}
		\end{minipage}
	}
	\subfigure[RCF ]{
		\begin{minipage}{0.12\linewidth}	
			\centering
			\includegraphics[width=2.2cm]{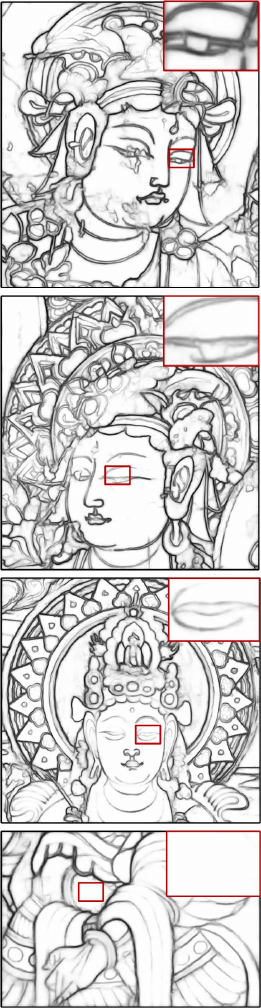}
		\end{minipage}
	}
	\subfigure[BDCN]{
		\begin{minipage}{0.12\linewidth}	
			\centering
			\includegraphics[width=2.2cm]{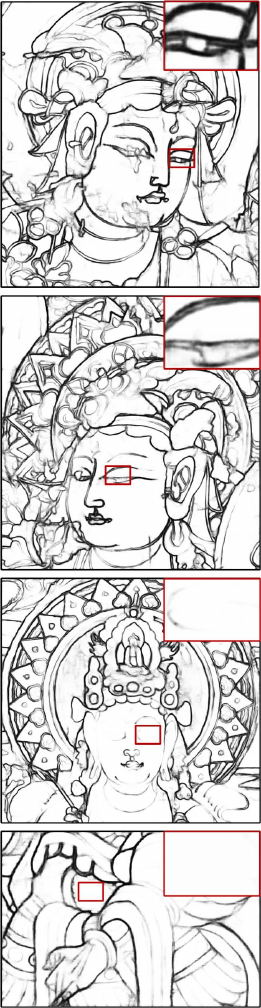}
		\end{minipage}
	}
	\subfigure[DEXINED]{
		\begin{minipage}{0.12\linewidth}	
			\centering
			\includegraphics[width=2.2cm]{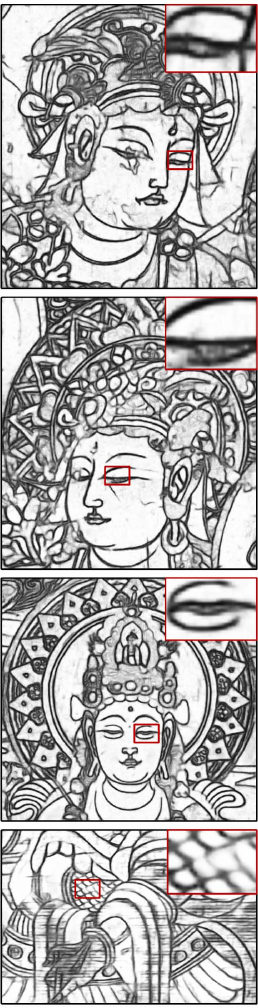}
		\end{minipage}
	}
    	\subfigure[RINDNet]{
		\begin{minipage}{0.12\linewidth}	
			\centering
			\includegraphics[width=2.2cm]{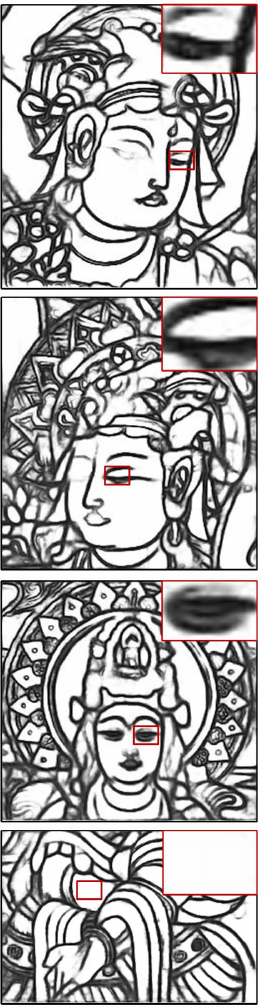}
		\end{minipage}
	}
	\subfigure[OUR]{
		\begin{minipage}{0.12\linewidth}	
			\centering
			\includegraphics[width=2.2cm]{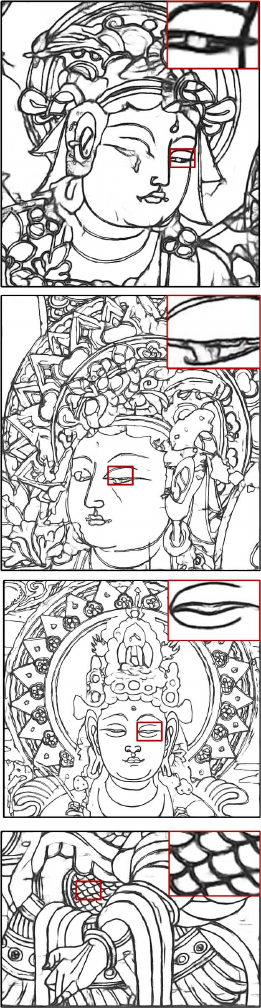}
		\end{minipage}
	}
	\hspace{0.1 cm}
	\caption{Visualization results of different algorithms on Dunhuang mural dataset}
	\label{fig:mural}
\end{figure*}

\begin{figure*}[h]% h ?????t?? ;b???p ???; ???[tbp]		
	\centering
	\subfigcapskip=5pt % ?????????????
	\subfigure[Input ]{
		\begin{minipage}{0.12\linewidth}	
			\centering
			\includegraphics[width=2.2cm]{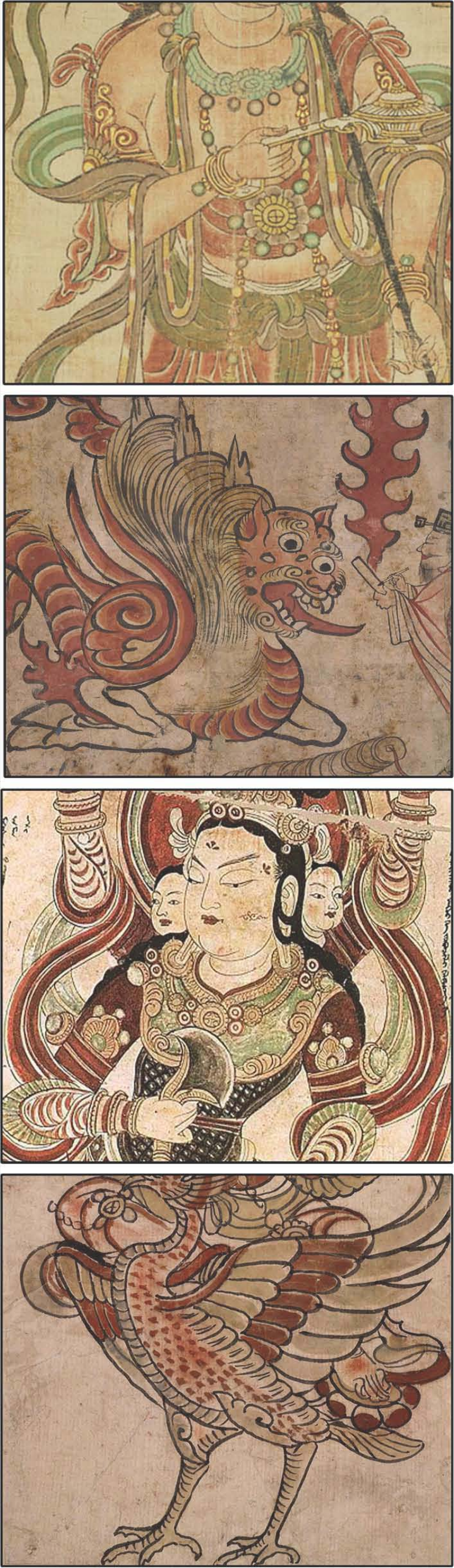}
		\end{minipage}
	}
	\subfigure[Canny]{
		\begin{minipage}{0.12\linewidth}	
			\centering
			\includegraphics[width=2.2cm]{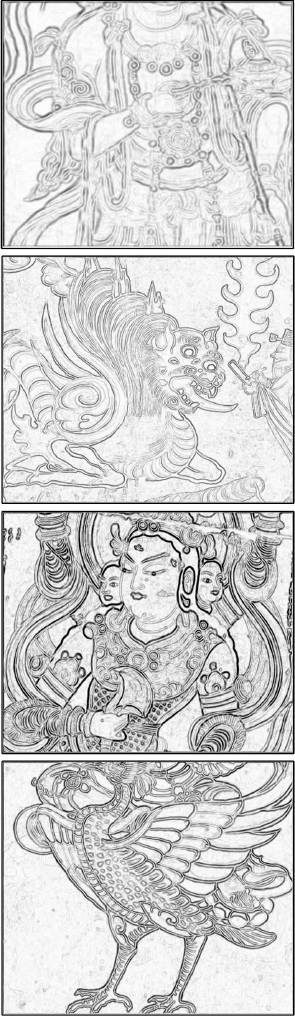}
		\end{minipage}
	}
	\subfigure[RCF ]{
		\begin{minipage}{0.12\linewidth}	
			\centering
			\includegraphics[width=2.2cm]{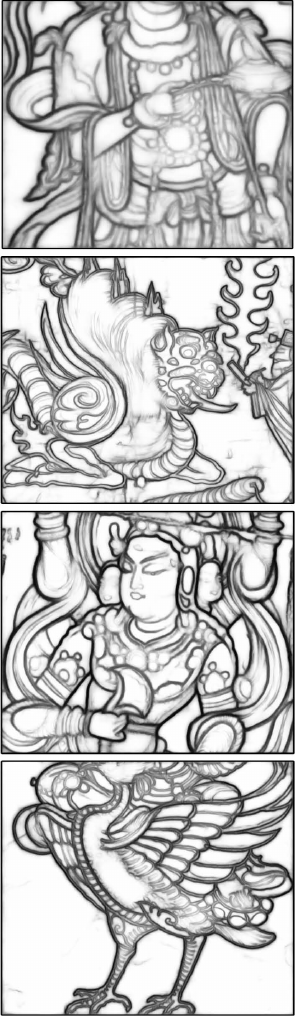}
		\end{minipage}
	}
	\subfigure[BDCN]{
		\begin{minipage}{0.12\linewidth}	
			\centering
			\includegraphics[width=2.2cm]{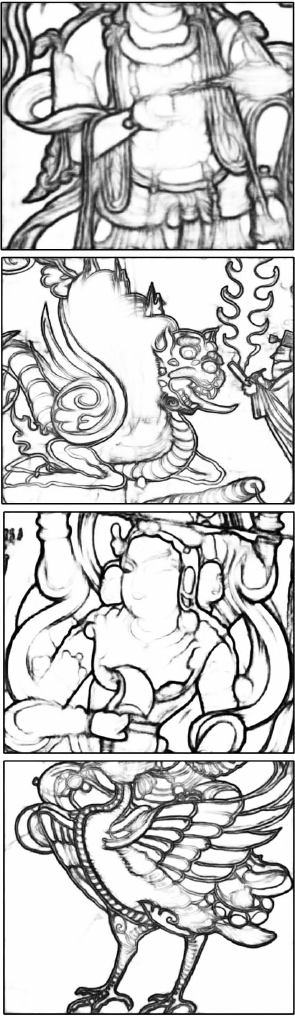}
		\end{minipage}
	}
	\subfigure[DEXINED]{
		\begin{minipage}{0.12\linewidth}	
			\centering
			\includegraphics[width=2.2cm]{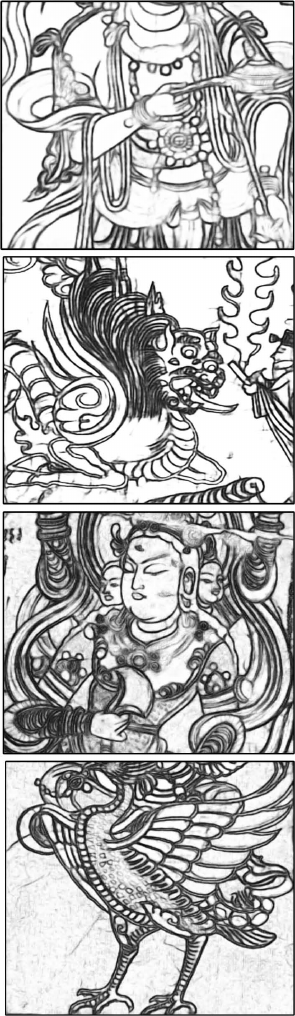}
		\end{minipage}
	}
    	\subfigure[RINDNet]{
		\begin{minipage}{0.12\linewidth}	
			\centering
			\includegraphics[width=2.2cm]{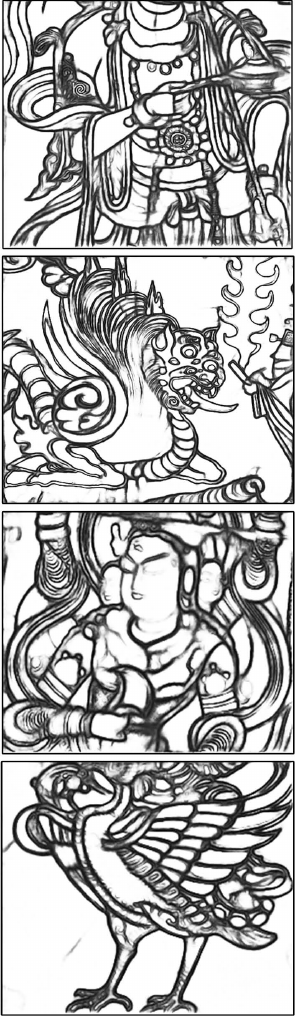}
		\end{minipage}
	}
	\subfigure[OUR]{
		\begin{minipage}{0.12\linewidth}	
			\centering
			\includegraphics[width=2.2cm]{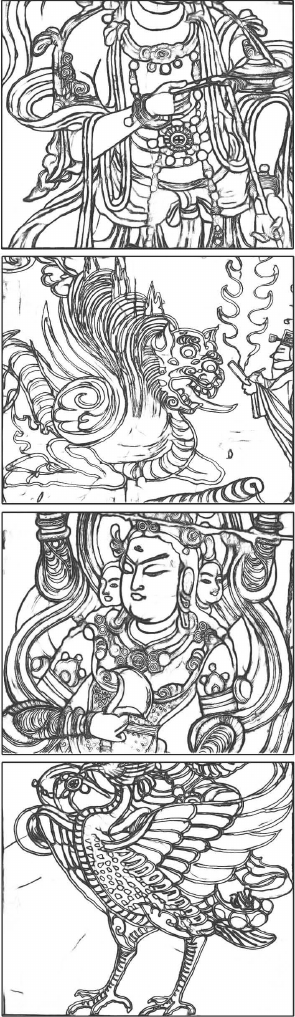}
		\end{minipage}
	}
	\hspace{0.1 cm}
	\caption{Visualization results on partial images of Dunhuang murals}
	\label{fig:mural2}
\end{figure*}

\begin{figure*}[h]% h ?????t?? ;b???p ???; ???[tbp]		
	\centering
	\subfigcapskip=5pt % ?????????????
	\subfigure[BSDS500]{
		\begin{minipage}{0.45\linewidth}	
			\centering
			\includegraphics[width=1\linewidth]{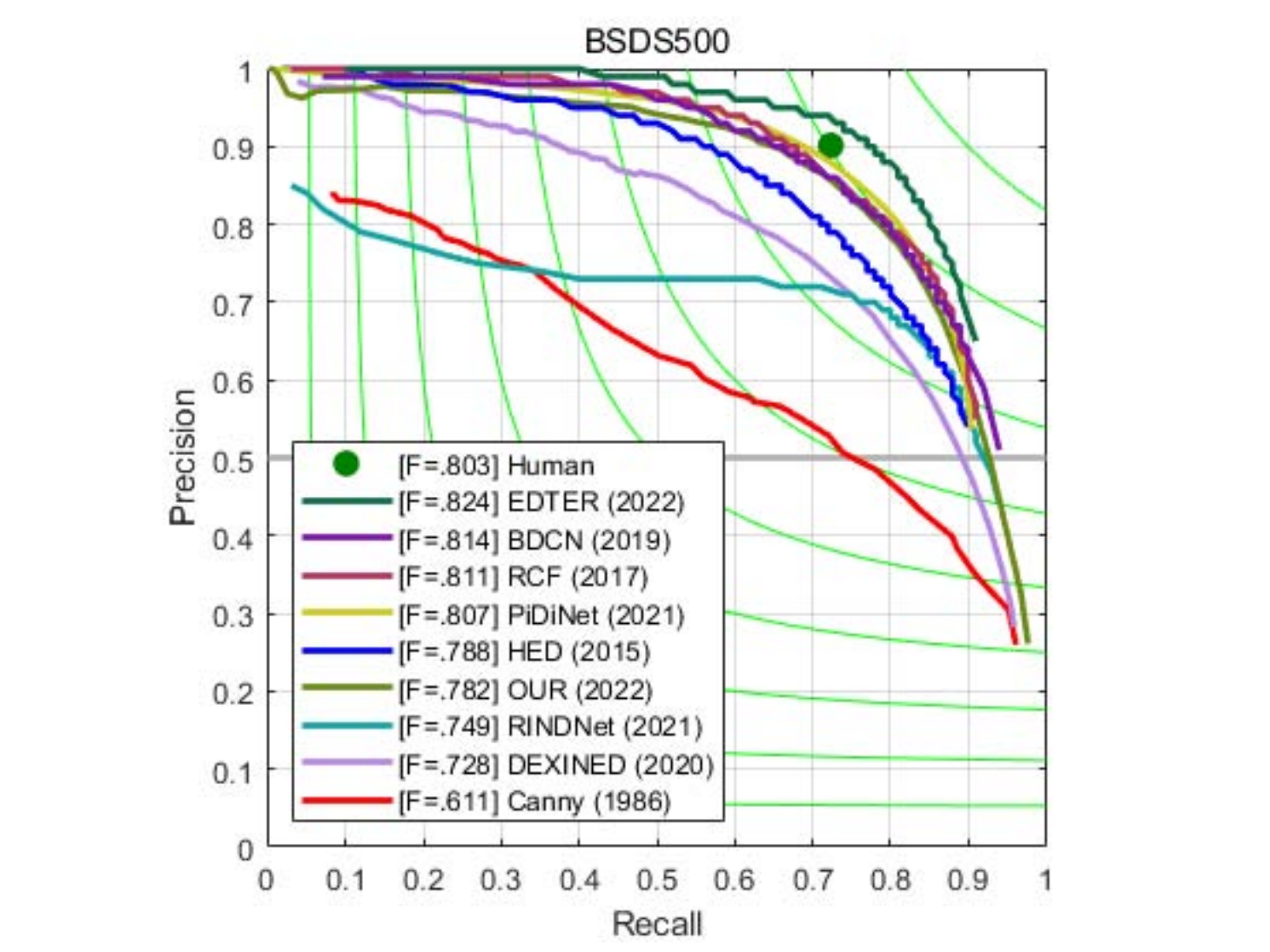}
		\end{minipage}
	}
	\subfigure[NYUD v2]{
		\begin{minipage}{0.45\linewidth}	
			\centering
			\includegraphics[width=1\linewidth]{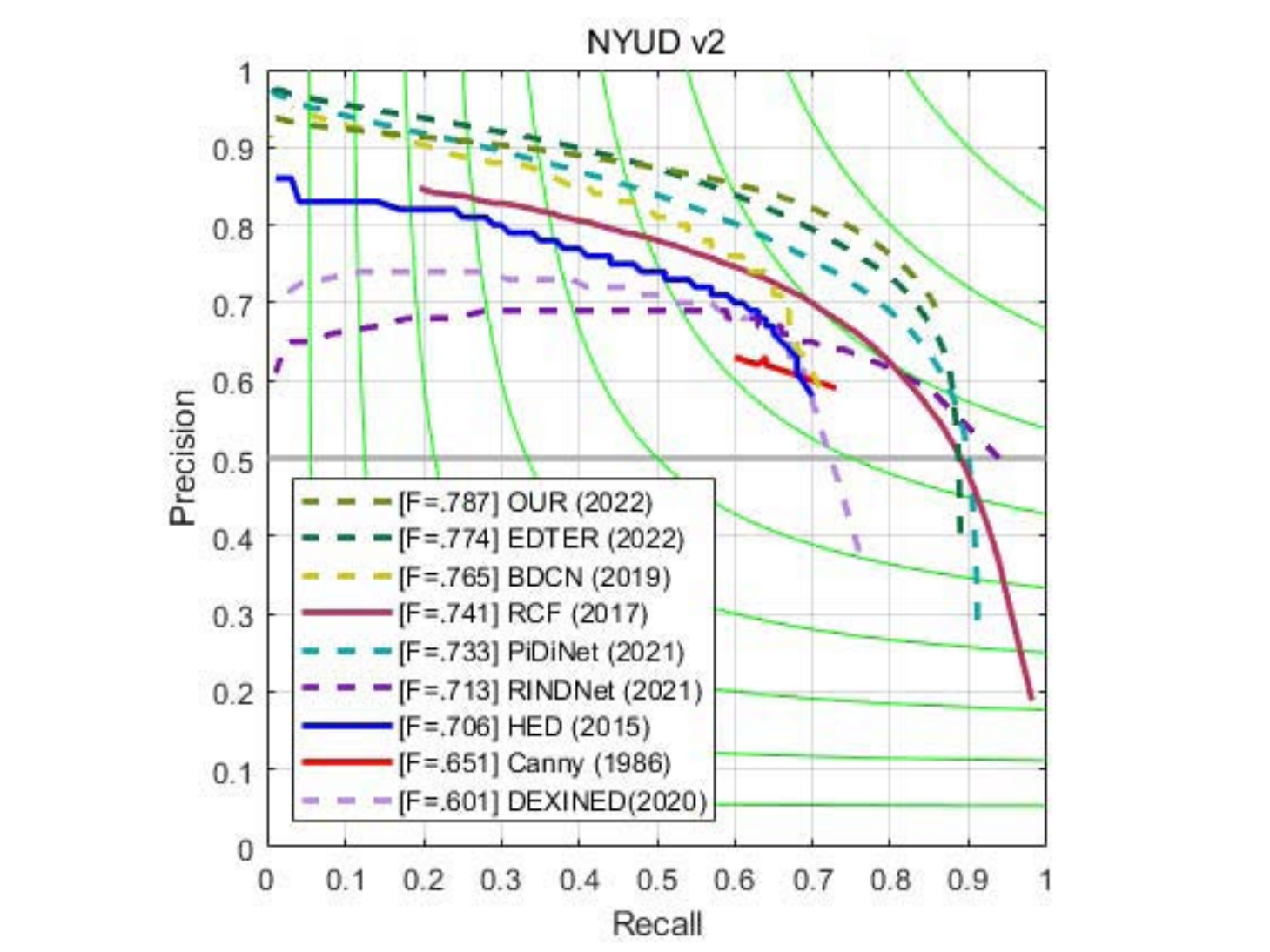}
		\end{minipage}
	}
~
    	\subfigure[Muticue]{
		\begin{minipage}{0.45\linewidth}	
			\centering
			\includegraphics[width=1\linewidth]{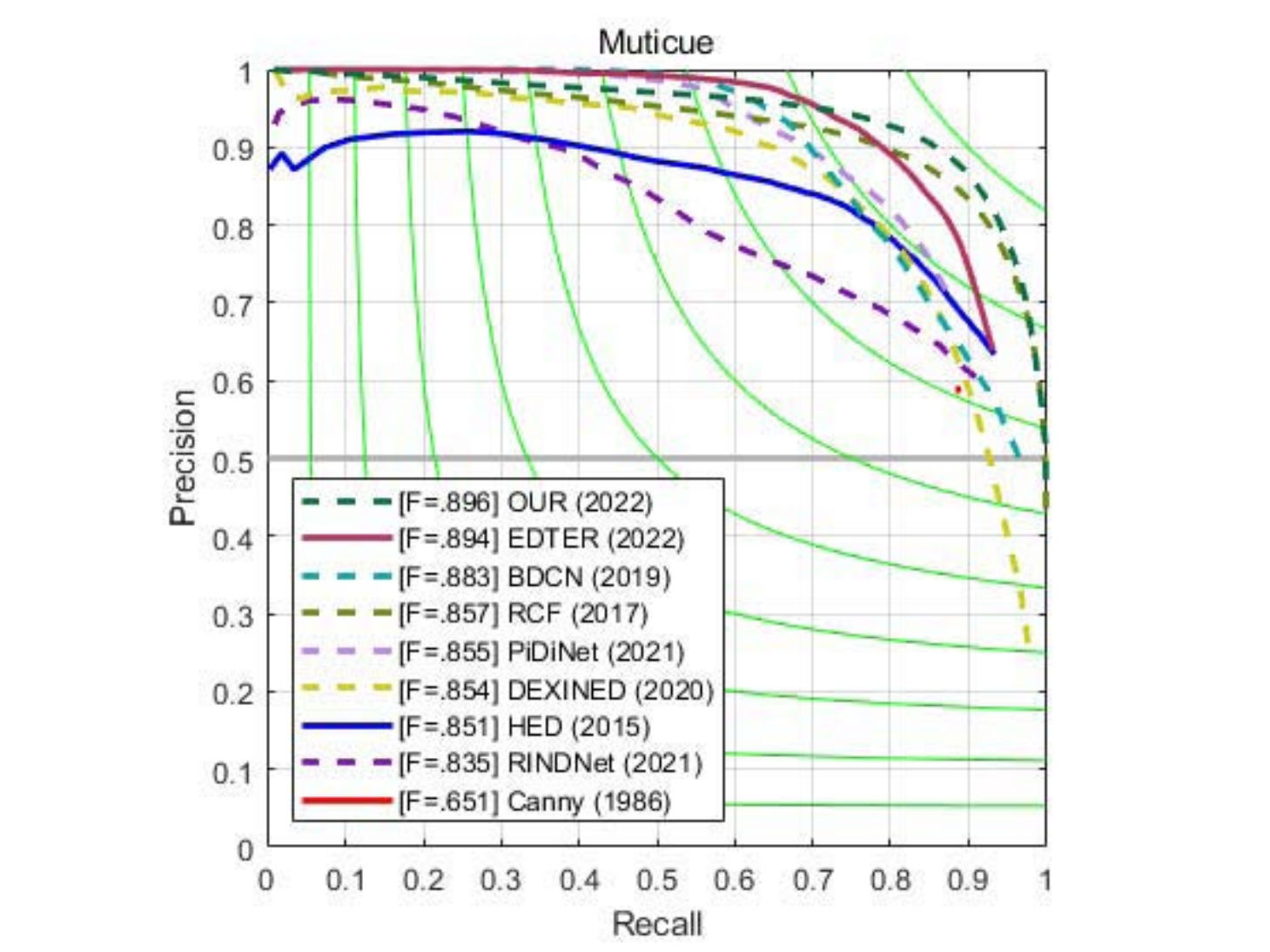}
		\end{minipage}
	}
	\subfigure[BIPED]{
		\begin{minipage}{0.45\linewidth}	
			\centering
			\includegraphics[width=1\linewidth]{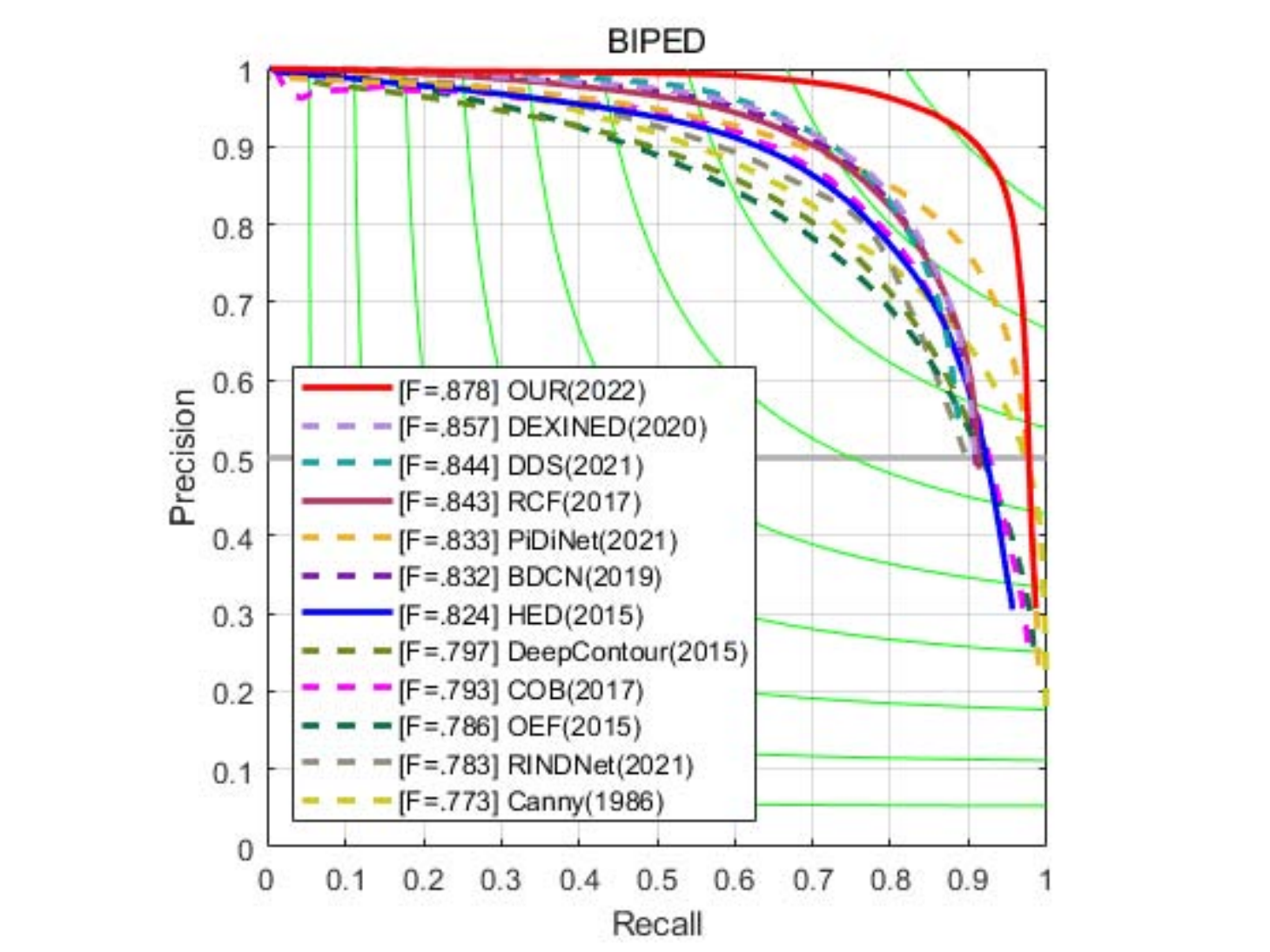}
		\end{minipage}
	}
	\hspace{0.1 cm}
	\caption{PR curves of different algorithms on the BSDS500, NYUD v2, Muticue, BIPED dataset}
	\label{fig:pr}
\end{figure*}

\begin{table}
\centering
\caption{Quantitative  results of BSDS500, NYUD v2, Multicue and BIPED dataset
and the state-o-the-art methods trained with the corresponding
datasets ( the $-$ indicates memory overflow on 2080Ti or the value cannot be calculated)}
\label{tab:all2}
\begin{tabular}{|c|ccc|ccc|}
\hline
\multirow{2}{*}{Methods} & \multicolumn{3}{c|}{\textbf{BSDS500}}            & \multicolumn{3}{c|}{\textbf{NYUD v2}}              \\
\cline{2-7}
                         & \textbf{ODS}   & \textbf{OIS}   & \textbf{AP}    & \textbf{ODS}   & \textbf{OIS}   & \textbf{AP}     \\
\hline
Canny(PAMI'86)           & 0.611          & 0.676          & 0.520          & 0.651          & 0.667          & 0.653           \\
HED(ICCV'15)             & 0.788          & 0.808          & 0.840          & 0.706          & 0.734          & 0.549           \\
RCF(CVPR'17)             & 0.811          & 0.830          & 0.846          & 0.741          & 0.757          & 0.749           \\
BDCN(CVPR'19)            & 0.814          & 0.833          & 0.847          & 0.765          & 0.780          & 0.760           \\
DexiNed(WACV'20)         & 0.728          & 0.745          & 0.689          & 0.601          & 0.614          & 0.485           \\
PiDiNet(CVPR'21)         & 0.807          & 0.823          & -              & 0.733          & 0.747          & -               \\
RINDNet(CVPR'21)         & 0.749          & 0.762          & 0.672          & 0.713          & 0.745          & 0.677           \\
EDTER(CVPR'22)           & \textbf{0.824} & \textbf{0.841} & \textbf{0.880} & 0.774          & 0.789          & 0.797           \\
OUR                      & 0.782          & 0.765          & 0.732          & \textbf{0.787} & \textbf{0.791} & \textbf{0.811}  \\
\hline
\multirow{2}{*}{Methods} & \multicolumn{3}{c|}{\textbf{Multicue}}           & \multicolumn{3}{c|}{\textbf{BIPED}}               \\
\cline{2-7}
                         & \textbf{ODS}   & \textbf{OIS}   & \textbf{AP}    & \textbf{ODS}   & \textbf{OIS}   & \textbf{AP}     \\
\hline
Canny(PAMI'86)           & 0.651          & 0.667          & 0.653          & 0.773          & 0.774          & 0.858           \\
HED(ICCV'15)             & 0.851          & 0.864          & -              & 0.824          & 0.845          & 0.865           \\
RCF(CVPR'17)             & 0.857          & 0.862          & -              & 0.843          & 0.854          & 0.881           \\
BDCN(CVPR'19)            & 0.883          & 0.896          & 0.933          & 0.832          & 0.853          & 0.887           \\
DexiNed(WACV'20)         & 0.854          & 0.862          & 0.915          & 0.857          & 0.861          & 0.904           \\
PiDiNet(CVPR'21)         & 0.855          & 0.860          & -              & 0.832          & 0.845          & 0.893           \\
RINDNet(CVPR'21)         & 0.835          & 0.843          & 0.896          & 0.783          & 0.784          & 0.760           \\
EDTER(CVPR'22)           & 0.894          & 0.897          & \textbf{0.942} & -              & -              & -               \\
OUR                      & \textbf{0.896} & \textbf{0.902} & 0.938          & \textbf{0.878} & \textbf{0.893} & \textbf{0.911}  \\
\hline
\end{tabular}
\end{table}

\textbf{\subsection{Analysis of results}}
%We have quantitatively and qualitatively verified the performance of the algorithm on multiple public datasets and Dunhuang mural datasets, and compared the experimental results with multiple edge detection algorithms. As can be seen from Table \ref{tab:all2}, our algorithm achieves better results in most algorithms. It can be seen from the three edge detection evaluation indicators ODS, OIS and AP values in the table that the model has improved by 1 to 2 percentage points.
%As can be seen from Figure \ref{fig:pr}, the PR curve of this model is significantly higher than that of other algorithms, indicating that our model is more accurate for the classification of edges and non-edges.
%In order to more intuitively show the edge detection effect of different algorithms, we visualize the edge detection results of different algorithms, and the results are shown in Figure \ref{fig:bsds500}, Figure \ref{fig:img_data}, and Figure \ref{fig:mural}. It can be seen from the whole visualization results that our model generates clearer and more meaningful edge maps, the main reason is that the combination of self-attention machine and convolution mechanism in the model can make better use of the above information of the image and local features.

We verify our model on four benchmark datasets for edge detection and Dunhuang mural dataset, and compared its experimental results quantitatively and qualitatively with several different edge detection algorithms.
From the three edge detection evaluation indicators $ODS$, $OIS$, and $AP$ values in Table \ref{tab:all2}, we can see that our algorithm has achieved the best results on the benchmark datasets NYUD v2, Multicue, and BIPED. Especially on the carefully annotated BIPED dataset, our method is obviously due to several other algorithms, and it can be seen from Figure~\ref{fig:pr}$(d)$ that the PR curve of our method is significantly higher than other algorithms, which indicates that our model can more accurately classify edge and non-edge samples.
For the mural dataset, our method generates clearer and more meaningful outlines of Buddha figures than other algorithms.
This is mainly due to the following three reasons:
\begin{enumerate}		
	\item The residual self-attention and convolution mixed module can make better use of global and local contextual information, so that the network can obtain the maximum receptive field information and fully integrate local information. This makes the generated edge map retain more meaningful edges.

	\item The cross-layer connection block in the dense network makes the feature information in the shallow layer in the network entirely merged with the feature information in the deep layer, which makes the generated edge map contain more detailed information.

    \item In this paper, the self-attention and convolution features are dynamically weighted and merged to ensure that the feature information in the two paths can only be  adaptively adjusted, which makes the local and global information widely used.

\end{enumerate}

\section{Conclusion}
\label{sec:conclusion}
%This paper proposes a new end-to-end edge detection framework by combining two paradigms of self-attention and convolution, and stacking different layers to capture long dependencies between contexts and extract effective features from feature maps. This hybrid paradigm combined with densely connected networks can effectively improve image edge detection results.
%Compared with existing edge detection algorithms, the experimental results show that our method achieves very competitive results both on the public dataset and on the Dunhuang mural dataset.

In this paper, we propose a method of edge detection based on the combination of self-attention and convolution to generate the line drawings of Dunhuang murals. The method is primarily divided into two parts: i) A new residual self-attention and convolution mixed module (Ramix) is proposed to merge the features extracted from the two paradigms of self-attention and convolution. ii) A new densely stacked edge detection network to efficiently transfer the feature information of two different feature extractors between the networks.
For the first time, the line drawing of Dunhuang murals is successfully realized by using an end-to-end edge detection network, and it is no longer necessary to perform non-maximum suppression processing on the line drawing generated in the next sequence.
The experimental results on public datasets and Dunhuang mural datasets show that our method can achieve results that are very competitive with existing methods.
The effectiveness of the proposed method is also demonstrated in terms of both quantitative and qualitative.
The model proposed in this paper does not require a pre-trained model, and the entire model can be trained from scratch. Since this model uses multiple self-attention paths and convolution fusion modules, and the fusion of the two paths uses a polymorphic weight adjustment mechanism, this will make the model require a longer training time. In future work, we will speed up the training of the entire model by providing prior information for the self-attention pathway.

%Future work will focus on the large-area restoration task of Dunhuang murals by combining the proposed edge detection method with an image inpainting network.\\

%\section*{Data availability statement}
%Data openly available in a public repository.
%\section*{Conflicts of interest}
%The authors declare that they have no conflict of interest.

\noindent \textbf{Conflicts of interest} The authors declare that they have no conflict of interest.\\

\noindent \textbf{Data availability statement} Data openly available in a public repository.

\textbf{\section*{Acknowledgment}}
This work was supported in part by the Outstanding Graduate ¡±Innovation Star¡± of Gansu Province(No.2022CXZX-202), the Gansu Provincial Department of Education University Teachers Innovation Fund Project (No.2023B-056),the Introduction of Talent Research Project of Northwest Minzu University (No.xbmuyjrc201904), and the Fundamental Research Funds for the Central Universities of Northwest Minzu University (No.31920220019, 31920220130), the Leading Talent of National Ethnic Affairs Commission (NEAC), the Young Talent of NEAC, and the Innovative Research Team of NEAC (2018) 98.

\bibliographystyle{spbasic_unsort}%??????? ?????? ????????? ??????????.bib?
\bibliography{XJ_ref}

\begin{thebibliography}{41}
\providecommand{\natexlab}[1]{#1}
\providecommand{\url}[1]{{#1}}
\providecommand{\urlprefix}{URL }
\expandafter\ifx\csname urlstyle\endcsname\relax
  \providecommand{\doi}[1]{DOI~\discretionary{}{}{}#1}\else
  \providecommand{\doi}{DOI~\discretionary{}{}{}\begingroup
  \urlstyle{rm}\Url}\fi
\providecommand{\eprint}[2][]{\url{#2}}

\bibitem[{Liu et~al.(2022)Liu, Du, Li, Wang, and Liu}]{liu2022dunhuang}
Liu B, Du S, Li J, Wang J, Liu W (2022) Dunhuang mural line drawing based on
  bi-dexined network and adaptive weight learning. In: Chinese Conference on
  Pattern Recognition and Computer Vision (PRCV), Springer, pp 279--292

\bibitem[{Marr and Hildreth(1980)}]{marr1980theory}
Marr D, Hildreth E (1980) Theory of edge detection. Proceedings of the Royal
  Society of London Series B Biological Sciences 207(1167):187--217

\bibitem[{Martin et~al.(2004)Martin, Fowlkes, and Malik}]{martin2004learning}
Martin DR, Fowlkes CC, Malik J (2004) Learning to detect natural image
  boundaries using local brightness, color, and texture cues. IEEE transactions
  on pattern analysis and machine intelligence 26(5):530--549

\bibitem[{Tariq~Jamal et~al.(2021)Tariq~Jamal, Ben~Ishak, and
  Abdel-Khalek}]{tariq2021tumor}
Tariq~Jamal A, Ben~Ishak A, Abdel-Khalek S (2021) Tumor edge detection in
  mammography images using quantum and machine learning approaches. Neural
  Computing and Applications 33(13):7773--7784

\bibitem[{Mohamed Ben~Ali(2021)}]{mohamed2021flexible}
Mohamed Ben~Ali Y (2021) Flexible edge detection and its enhancement by smell
  bees optimization algorithm. Neural Computing and Applications
  33(16):10021--10041

\bibitem[{Ziou et~al.(1998)Ziou, Tabbone et~al.}]{ziou1998edge}
Ziou D, Tabbone S, et~al. (1998) Edge detection techniques-an overview. Pattern
  Recognition and Image Analysis C/C of Raspoznavaniye Obrazov I Analiz
  Izobrazhenii 8:537--559

\bibitem[{Dollar et~al.(2006)Dollar, Tu, and Belongie}]{dollar2006supervised}
Dollar P, Tu Z, Belongie S (2006) Supervised learning of edges and object
  boundaries. In: 2006 IEEE Computer Society Conference on Computer Vision and
  Pattern Recognition (CVPR'06), IEEE, vol~2, pp 1964--1971

\bibitem[{Pu et~al.(2022)Pu, Huang, Liu, Guan, and Ling}]{pu2022edter}
Pu M, Huang Y, Liu Y, Guan Q, Ling H (2022) Edter: Edge detection with
  transformer. In: Proceedings of the IEEE/CVF Conference on Computer Vision
  and Pattern Recognition, pp 1402--1412

\bibitem[{Pan et~al.(2022)Pan, Ge, Lu, Song, Chen, Huang, and
  Huang}]{pan2022integration}
Pan X, Ge C, Lu R, Song S, Chen G, Huang Z, Huang G (2022) On the integration
  of self-attention and convolution. In: Proceedings of the IEEE/CVF Conference
  on Computer Vision and Pattern Recognition, pp 815--825

\bibitem[{Srinivas et~al.(2021)Srinivas, Lin, Parmar, Shlens, Abbeel, and
  Vaswani}]{srinivas2021bottleneck}
Srinivas A, Lin TY, Parmar N, Shlens J, Abbeel P, Vaswani A (2021) Bottleneck
  transformers for visual recognition. In: Proceedings of the IEEE/CVF
  conference on computer vision and pattern recognition, pp 16519--16529

\bibitem[{Canny(1986)}]{canny1986computational}
Canny J (1986) A computational approach to edge detection. IEEE Transactions on
  pattern analysis and machine intelligence (6):679--698

\bibitem[{Kanopoulos et~al.(1988)Kanopoulos, Vasanthavada, and
  Baker}]{kanopoulos1988design}
Kanopoulos N, Vasanthavada N, Baker RL (1988) Design of an image edge detection
  filter using the sobel operator. IEEE Journal of solid-state circuits
  23(2):358--367

\bibitem[{Chaple et~al.(2015)Chaple, Daruwala, and
  Gofane}]{chaple2015comparisions}
Chaple GN, Daruwala R, Gofane MS (2015) Comparisions of robert, prewitt, sobel
  operator based edge detection methods for real time uses on fpga. In: 2015
  International Conference on Technologies for Sustainable Development (ICTSD),
  IEEE, pp 1--4

\bibitem[{Lin et~al.(2022)Lin, Zhang, and Hu}]{lin2022bio}
Lin C, Zhang Z, Hu Y (2022) Bio-inspired feature enhancement network for edge
  detection. Applied Intelligence pp 1--16

\bibitem[{Al-Amaren et~al.(2022)Al-Amaren, Ahmad, and Swamy}]{al2022low}
Al-Amaren A, Ahmad MO, Swamy M (2022) A low-complexity residual deep neural
  network for image edge detection. Applied Intelligence pp 1--18

\bibitem[{Liang and Liu(2021)}]{liang2021coarse}
Liang D, Liu X (2021) Coarse-to-fine foreground segmentation based on
  co-occurrence pixel-block and spatio-temporal attention model. In: 2020 25th
  International Conference on Pattern Recognition (ICPR), IEEE, pp 3807--3813

\bibitem[{Deng et~al.(2018)Deng, Shen, Liu, Wang, and Liu}]{deng2018learning}
Deng R, Shen C, Liu S, Wang H, Liu X (2018) Learning to predict crisp
  boundaries. In: Proceedings of the European Conference on Computer Vision
  (ECCV), pp 562--578

\bibitem[{He et~al.(2019)He, Zhang, Yang, Shan, and Huang}]{he2019bi}
He J, Zhang S, Yang M, Shan Y, Huang T (2019) Bi-directional cascade network
  for perceptual edge detection. In: Proceedings of the IEEE/CVF Conference on
  Computer Vision and Pattern Recognition, pp 3828--3837

\bibitem[{Hu et~al.(2019)Hu, Chen, Li, and Feng}]{hu2019dynamic}
Hu Y, Chen Y, Li X, Feng J (2019) Dynamic feature fusion for semantic edge
  detection. arXiv preprint arXiv:190209104

\bibitem[{Huang et~al.(2017)Huang, Liu, Van Der~Maaten, and
  Weinberger}]{huang2017densely}
Huang G, Liu Z, Van Der~Maaten L, Weinberger KQ (2017) Densely connected
  convolutional networks. In: Proceedings of the IEEE conference on computer
  vision and pattern recognition, pp 4700--4708

\bibitem[{Konishi et~al.(2003)Konishi, Yuille, Coughlan, and
  Zhu}]{konishi2003statistical}
Konishi S, Yuille AL, Coughlan JM, Zhu SC (2003) Statistical edge detection:
  Learning and evaluating edge cues. IEEE Transactions on Pattern Analysis and
  Machine Intelligence 25(1):57--74

\bibitem[{Xie and Tu(2015)}]{xie2015holistically}
Xie S, Tu Z (2015) Holistically-nested edge detection. In: Proceedings of the
  IEEE international conference on computer vision, pp 1395--1403

\bibitem[{Liu et~al.(2017)Liu, Cheng, Hu, Wang, and Bai}]{liu2017richer}
Liu Y, Cheng MM, Hu X, Wang K, Bai X (2017) Richer convolutional features for
  edge detection. In: Proceedings of the IEEE conference on computer vision and
  pattern recognition, pp 3000--3009

\bibitem[{Poma et~al.(2020)Poma, Riba, and Sappa}]{poma2020dense}
Poma XS, Riba E, Sappa A (2020) Dense extreme inception network: Towards a
  robust cnn model for edge detection. In: Proceedings of the IEEE/CVF Winter
  Conference on Applications of Computer Vision, pp 1923--1932

\bibitem[{Su et~al.(2021)Su, Liu, Yu, Hu, Liao, Tian, Pietik{\"a}inen, and
  Liu}]{su2021pixel}
Su Z, Liu W, Yu Z, Hu D, Liao Q, Tian Q, Pietik{\"a}inen M, Liu L (2021) Pixel
  difference networks for efficient edge detection. In: Proceedings of the
  IEEE/CVF International Conference on Computer Vision, pp 5117--5127

\bibitem[{Iandola et~al.(2014)Iandola, Moskewicz, Karayev, Girshick, Darrell,
  and Keutzer}]{iandola2014densenet}
Iandola F, Moskewicz M, Karayev S, Girshick R, Darrell T, Keutzer K (2014)
  Densenet: Implementing efficient convnet descriptor pyramids. arXiv preprint
  arXiv:14041869

\bibitem[{Deng and Liu(2020)}]{deng2020deep}
Deng R, Liu S (2020) Deep structural contour detection. In: Proceedings of the
  28th ACM international conference on multimedia, pp 304--312

\bibitem[{Shen et~al.(2015)Shen, Wang, Wang, Bai, and
  Zhang}]{shen2015deepcontour}
Shen W, Wang X, Wang Y, Bai X, Zhang Z (2015) Deepcontour: A deep convolutional
  feature learned by positive-sharing loss for contour detection. In:
  Proceedings of the IEEE conference on computer vision and pattern
  recognition, pp 3982--3991

\bibitem[{Yang et~al.(2015)Yang, Gao, Guo, Li, and Li}]{yang2015boundary}
Yang KF, Gao SB, Guo CF, Li CY, Li YJ (2015) Boundary detection using
  double-opponency and spatial sparseness constraint. IEEE Transactions on
  Image Processing 24(8):2565--2578

\bibitem[{Dosovitskiy et~al.(2020)Dosovitskiy, Beyer, Kolesnikov, Weissenborn,
  Zhai, Unterthiner, Dehghani, Minderer, Heigold, Gelly
  et~al.}]{dosovitskiy2020image}
Dosovitskiy A, Beyer L, Kolesnikov A, Weissenborn D, Zhai X, Unterthiner T,
  Dehghani M, Minderer M, Heigold G, Gelly S, et~al. (2020) An image is worth
  16x16 words: Transformers for image recognition at scale. arXiv preprint
  arXiv:201011929

\bibitem[{Yu et~al.(2018)Yu, Liu, Zou, Feng, Ramalingam, Kumar, and
  Kautz}]{yu2018simultaneous}
Yu Z, Liu W, Zou Y, Feng C, Ramalingam S, Kumar B, Kautz J (2018) Simultaneous
  edge alignment and learning. In: Proceedings of the European Conference on
  Computer Vision (ECCV), pp 388--404

\bibitem[{Fu et~al.(2019)Fu, Liu, Tian, Li, Bao, Fang, and Lu}]{fu2019dual}
Fu J, Liu J, Tian H, Li Y, Bao Y, Fang Z, Lu H (2019) Dual attention network
  for scene segmentation. In: Proceedings of the IEEE/CVF conference on
  computer vision and pattern recognition, pp 3146--3154

\bibitem[{Acuna et~al.(2019)Acuna, Kar, and Fidler}]{acuna2019devil}
Acuna D, Kar A, Fidler S (2019) Devil is in the edges: Learning semantic
  boundaries from noisy annotations. In: Proceedings of the IEEE/CVF Conference
  on Computer Vision and Pattern Recognition, pp 11075--11083

\bibitem[{Xu et~al.(2017)Xu, Ouyang, Alameda-Pineda, Ricci, Wang, and
  Sebe}]{xu2017learning}
Xu D, Ouyang W, Alameda-Pineda X, Ricci E, Wang X, Sebe N (2017) Learning deep
  structured multi-scale features using attention-gated crfs for contour
  prediction. Advances in neural information processing systems 30

\bibitem[{Zhang et~al.(2016)Zhang, Xing, Shi, and Yang}]{zhang2016semicontour}
Zhang Z, Xing F, Shi X, Yang L (2016) Semicontour: A semi-supervised learning
  approach for contour detection. In: Proceedings of the IEEE Conference on
  Computer Vision and Pattern Recognition, pp 251--259

\bibitem[{Yu et~al.(2017)Yu, Feng, Liu, and Ramalingam}]{yu2017casenet}
Yu Z, Feng C, Liu MY, Ramalingam S (2017) Casenet: Deep category-aware semantic
  edge detection. In: Proceedings of the IEEE conference on computer vision and
  pattern recognition, pp 5964--5973

\bibitem[{Chen et~al.(2022)Chen, Wu, Wang, Hu, Hu, Ding, Cheng, and
  Wang}]{chen2022mixformer}
Chen Q, Wu Q, Wang J, Hu Q, Hu T, Ding E, Cheng J, Wang J (2022) Mixformer:
  Mixing features across windows and dimensions. In: Proceedings of the
  IEEE/CVF Conference on Computer Vision and Pattern Recognition, pp 5249--5259

\bibitem[{Zheng et~al.(2021)Zheng, Lu, Zhao, Zhu, Luo, Wang, Fu, Feng, Xiang,
  Torr et~al.}]{zheng2021rethinking}
Zheng S, Lu J, Zhao H, Zhu X, Luo Z, Wang Y, Fu Y, Feng J, Xiang T, Torr PH,
  et~al. (2021) Rethinking semantic segmentation from a sequence-to-sequence
  perspective with transformers. In: Proceedings of the IEEE/CVF conference on
  computer vision and pattern recognition, pp 6881--6890

\bibitem[{Zhu et~al.(2020)Zhu, Su, Lu, Li, Wang, and Dai}]{zhu2020deformable}
Zhu X, Su W, Lu L, Li B, Wang X, Dai J (2020) Deformable detr: Deformable
  transformers for end-to-end object detection. arXiv preprint arXiv:201004159

\bibitem[{Zhao et~al.(2020)Zhao, Jia, and Koltun}]{zhao2020exploring}
Zhao H, Jia J, Koltun V (2020) Exploring self-attention for image recognition.
  In: Proceedings of the IEEE/CVF Conference on Computer Vision and Pattern
  Recognition, pp 10076--10085

\bibitem[{Pu et~al.(2021)Pu, Huang, Guan, and Ling}]{pu2021rindnet}
Pu M, Huang Y, Guan Q, Ling H (2021) Rindnet: Edge detection for discontinuity
  in reflectance, illumination, normal and depth. In: Proceedings of the
  IEEE/CVF International Conference on Computer Vision, pp 6879--6888

\end{thebibliography}
\end{document}